\documentclass[final,5p,times,twocolumn]{elsarticle}

\usepackage{amssymb}

\usepackage{xcolor}

\usepackage{amsmath}
\usepackage{siunitx}

\usepackage{ifthen}
\newboolean{final}
\setboolean{final}{true}

\ifthenelse{\boolean{final}}
{
    \newcommand{\new}[1]{#1}
    \newcommand{\remove}[1]{}
    \newcommand{\replace}[2]{#2}
}{
    \usepackage{color,soul}
    \DeclareRobustCommand{\new}[1]{{\sethlcolor{yellow}\hl{#1}}}
    \DeclareRobustCommand{\remove}[1]{{\sethlcolor{red}\hl{#1}}}
    \DeclareRobustCommand{\replace}[2]{{\sethlcolor{red}\hl{#1}}{\sethlcolor{yellow}\hl{#2}}}
}

\usepackage{url}

\RequirePackage[format=plain,labelformat=simple,labelsep=period,font=small,compatibility=false]{caption}
\RequirePackage[font=footnotesize,skip=3pt,subrefformat=parens]{subcaption}

\journal{Microprocessors and Microsystems}

\begin{document}

\begin{frontmatter}



\title{High Throughput Event Filtering: The Interpolation-based DIF Algorithm Hardware Architecture}


\author[inst1]{Marcin Kowalczyk}

\author[inst1]{Tomasz Kryjak}

\affiliation[inst1]{organization={Embedded Vision Systems Group, Department of Automatic Control and Robotics, Faculty of Electrical Engineering, Automatics, Computer Science and Biomedical Engineering AGH, University of Krakow},
            country={Poland}}

\begin{abstract}
In recent years, there has been rapid development in the field of event vision. It manifests itself both on the technical side, as better and better event sensors are available, and on the algorithmic side, as more and more applications of this technology are proposed and scientific papers are published.
However, the data stream from these sensors typically contains a significant amount of noise, which varies depending on factors such as the degree of illumination in the observed scene or the temperature of the sensor.
We propose a hardware architecture of the Distance-based Interpolation with Frequency \new{Weights }(DIF) filter and implement it on an FPGA chip.
To evaluate the algorithm and compare it with other solutions, we have prepared a new high-resolution event dataset, which we are also releasing to the community.
Our architecture achieved a throughput of \replace{445.83}{403.39} million events per second (MEPS) for a sensor resolution of \(1280 \times 720\) and \replace{469.04}{428.45} MEPS for a resolution of \(640 \times 480\). The average\new{ values of the} Area Under the Receiver Operating Characteristic (AUROC) index\remove{ values} ranged from 0.844 to 0.999, depending on the dataset, which is comparable to the state-of-the-art filtering solutions\new{,} but with much higher throughput and better operation over a wide range of noise levels.
\end{abstract}


\end{frontmatter}


\section{Introduction}
\label{sec:introduction}
\begin{figure}
    \centering
    \includegraphics[width=0.8\linewidth]{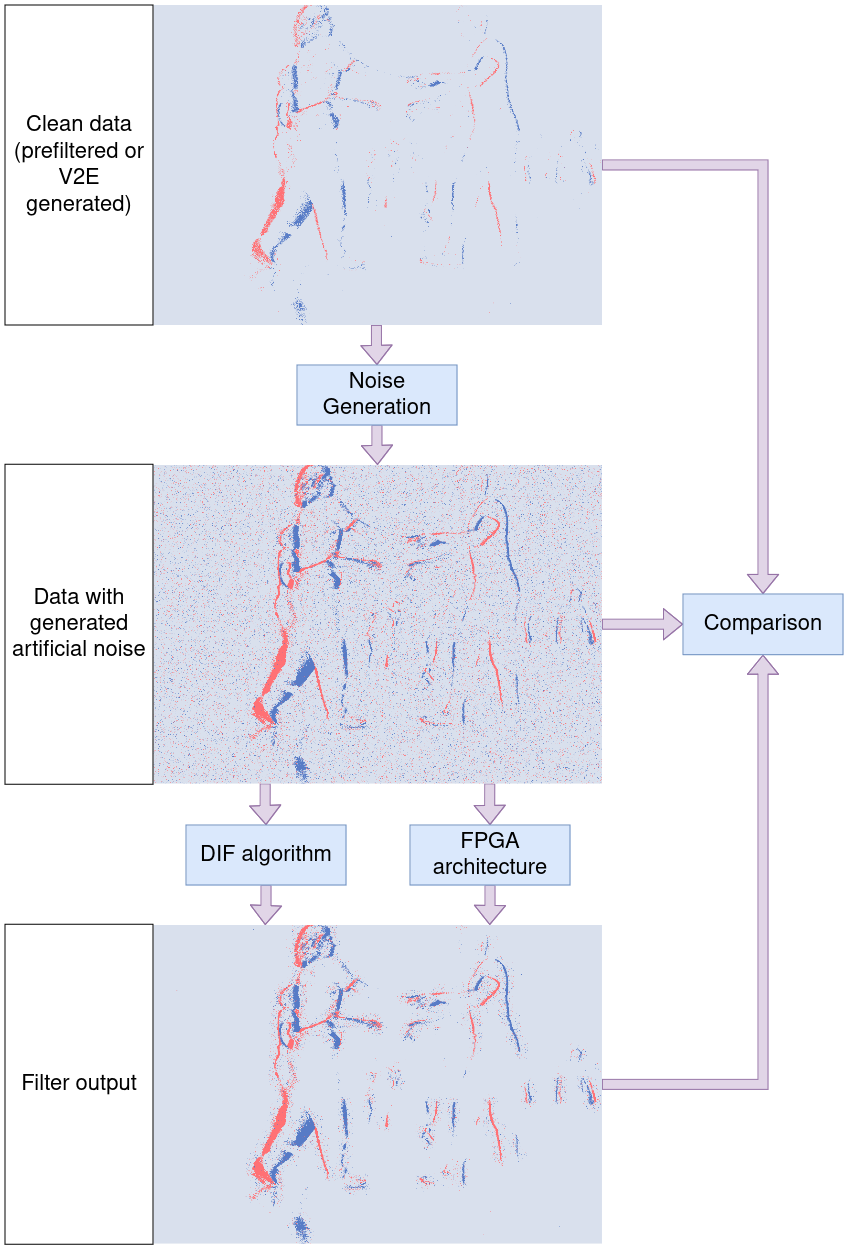}
    \caption{
    A diagram of the proposed event filtering system. Marked noise was added to the input data, which were then filtered using the proposed algorithms. The effectiveness of the filtering was evaluated based on the input and output data.}
    \label{fig:Introduction}
\end{figure}

Event cameras, also referred to as neuromorphic cameras or dynamic vision sensors\new{ (DVS)}, are a novel type of vision sensor inspired by biological systems. They detect luminance changes asynchronously and with high temporal resolution.
Unlike traditional frame-based cameras, event cameras respond to changes in pixel brightness with microsecond precision instead of recording the absolute value of the intensity and colour of the light incident on the matrix pixels.
This approach has a number of advantages, but it also has some disadvantages. These are discussed in detail in Section \ref{sec:preliminaries}.
The aim of this work is to develop an event data filtering algorithm that is highly efficient regardless of the amount of noise and requires a small amount of memory. This will enable the implementation in the resources of an \replace{FPGA}{Field Programmable Gate Array (FPGA)} chip.
The algorithm's effectiveness is measured by its ability to remove noise while retaining relevant events related to movement or brightness changes in the observed scene.
Figure \ref{fig:Introduction} shows a simplified schematic of the proposed system. Marked noise was intentionally added to the input data, which were then filtered using the proposed algorithms. The filter effectiveness was determined \replace{based on}{on the basis of} the comparison of input and output data.

The topic of event data filtering has been extensively discussed in the scientific literature, as detailed in Section \ref{sec:previous}. However, these studies often neglect to address the throughput and speed of the proposed solutions.
Several papers have addressed the issue of event stream filtering using FPGA computational resources. Most of these architectures were designed for sensors with a resolution much smaller than \(1280 \times 720\) pixels. However, the use of FPGAs in the field of neuromorphic vision is promising and increasingly popular due to the possibility of pipeline processing of the event stream.

\subsection*{Contributions}
The research described in our publication \mbox{\cite{Kowalczyk_2023_CVPR}} served as the starting point for this work. It has been extended to include in-depth testing of the proposed algorithms, analysis in terms of matching to FPGA resources, a proposal for hardware implementation on an FPGA chip of one of the solutions, and evaluation of its effectiveness.
The main contribution of this paper can be summarised as follows.
\begin{itemize}
    \item A novel custom event stream filtering algorithm \replace{DIF (Distance-based Interpolation with Frequency weights)}{Distance-based Interpolation with Frequency Weights (DIF)} developed using the hardware-aware design methodology. It has been designed to maximise the feasibility \replace{in}{of} FPGA computational resources, achieving the highest possible throughput while maintaining minimal resource utilisation and good filtering performance. The module was verified in hardware, in FPGA logic.
    \item A new publicly available \(1280 \times 720\) resolution dataset to evaluate the effectiveness of filtering algorithms for DVS.
    \item An in-depth evaluation of the \replace{Distance Based Interpolation with Frequency Weights (DIF)}{DIF} algorithm \replace{as well as}{and} comparison with other approaches.
\end{itemize}

To the best knowledge of the authors, this is the first event processing hardware architecture that connects efficiency comparable to \replace{the SOTA}{state-of-the-art (SOTA)} methods with throughput of \replace{over}{more than} 400 million events per second for a resolution of \(1280 \times 720\) pixels.

The rest of the article is structured as follows.
Section \ref{sec:preliminaries} presents basic information about event cameras.
Section \ref{sec:previous} analyses articles related to event data filtering.
Section \ref{sec:algorithms} describes the operation methods of the filtering algorithms considered.
Section \ref{sec:datasets} presents the datasets used to evaluate the effectiveness of filtering.
Section \ref{sec:evaluation} presents the results of the algorithms on the datasets.
Section \ref{sec:hardwarearchitecture} discusses the hardware\new{ architecture} realising the proposed method.
Section \ref{sec:comparison} provides a comparison of the designed solution with state-of-the-art approaches.
Finally, Section \ref{sec:summary} summarises the work that has been done and presents conclusions.

\section{Preliminaries}
\label{sec:preliminaries}
Event cameras are a novel type of video sensor that takes inspiration from the natural mechanisms observed in the human eye. These devices have a unique feature where each pixel on the sensor reacts independently to changes in ambient light. The key element is how the pixels register the change in the logarithm of light intensity. The process is described by a threshold mechanism that determines whether the difference in light intensity for a given pixel has exceeded a predetermined threshold \(C\), as expressed by \remove{Equation }\eqref{eq:generate}.

\begin{equation}
L(u_i,t_i) - L(u_i, t_i - \Delta t_i) \geq p_i C,
\label{eq:generate}
\end{equation}
where \(L(u, t)\) is the logarithm of the light intensity for a pixel with location \(u = (x, y)\) at time \(t\), \(\Delta t\) is the elapsed time since the last recorded event for this pixel.

The resulting event stream can be described as a sequence of values \(E = \{x, y, t, p\}\), where \(p\) defines the polarisation of the light change. This characteristic enables data to be recorded only when there are changes in brightness recorded by the pixels. In scenarios where the scene remains static, no data is generated, resulting in significant energy savings.

This method of recording vision data offers several advantages.
Firstly, it provides high temporal resolution as it monitors intensity changes at up to 1 MHz.
This enables the observation of even very fast movements with minimal blurring, which is a common issue with frame cameras.
\replace{Additionally}{In addition}, it \replace{boasts}{has} a large dynamic range of over 120 dB. This is significantly higher than that of even high-quality conventional cameras.
The sensor also has a low latency, which is the time between a change in the observed scene and the sending of the corresponding event from the sensor. The independent operation of the pixels eliminates the need to wait for the acquisition of the entire frame.
Finally, these sensors have low average power consumption. They only transmit brightness change information, which reduces data redundancy. However, it is important to note that the power consumption and data generated are heavily dependent on the dynamics of the observed scene.
The sensor itself can consume up to several hundred mW \cite{gallego2020event}.

Processing event camera data presents a significant challenge due to its asynchronous nature and unique characteristics. This requires the development of specialised processing techniques tailored to exploit data in the form of a sparse point cloud in three-dimensional space.
Moreover, conventional image processing algorithms cannot be directly applied to this type of data.
There is no available information on the absolute intensity and colour of the light recorded for a specific pixel.
Event cameras are susceptible to noise due to the non-ideal nature of the sensors, shot noise of the photons, and noise from the analogue part of the pixel array.
This type of noise appears as events\new{ that are} unrelated to objects, camera movements, or lighting changes.
It is relatively easy to identify, as it has the form of ``impulse noise'' (salt and pepper).
This noise can negatively impact the performance of classification, detection, or segmentation algorithms, leading to increased power consumption and computing power usage.
The number of noisy events at the output of a sensor depends on several factors. One of them is the illumination level of the observed scene. The brighter the scene, the less noise is recorded. In very dark scenes, noise can account for a significant proportion of the recorded data. Another factor is the\new{ sensor} temperature \cite{nozaki2017temperature}. Therefore, a different noise intensity may be recorded immediately after the sensor is switched on than after some time when its temperature rises. In addition to this, sensor performance parameters, such as comparator thresholds, i.e. the variable \(C\) in \eqref{eq:generate}, are also very important.
The excellent paper \cite{gallego2020event} provides a comprehensive explanation of the functioning of the event camera, as well as a~in-depth discussion on possible applications.

\section{Previous work}
\label{sec:previous}
The scientific literature has covered the topic of event visual data filtering many times due to its practical importance.
This chapter offers an overview of the various approaches that have been taken to address this issue.

In 2008, a paper \cite{delbruck2008frame} introduced a method to remove uncorrelated background activity.
The method involved deleting events that had no recorded activity in their surroundings within a preset time window.
To implement this method, a map of timestamps with a size equal to the\new{ sensor} resolution multiplied by 2 needed to be stored in memory. A DVS sensor with a resolution of \(128 \times 128\) was used. This method is widely known as Nearest-Neighbour filtering.
The paper proposed a filtering method that could process up to 10 \replace{MEPS}{million events per second (MEPS)} on a 2005 laptop.

The paper \cite{linares2015usb3} presented a framework for removing uncorrelated noise on an FPGA platform. The processed events' timestamps were stored in a \(128 \times 128\) register matrix.
For each input event, the previously stored timestamp was read from the matrix and the difference between the input and the read \replace{one}{timestamp} was calculated. If the difference exceeded a preset threshold, the event was deleted and a new timestamp was written \replace{into}{at} the same location in the array.
The paper does not provide information on the maximum operating frequency of the proposed architecture or its throughput.

The paper \cite{liu2015design} proposed a background activity filter that allowed only correlated events in space-time to be processed further.
The aim was to reduce communication and computational overhead while increasing the rate of correct information. The solution was designed using a \(128 \times 128\) matrix chip with \SI{20}{\um} \(\times\) \SI{20}{\um} cells.
Each cell combines spatial subsampling with a time window based on current integration.
The designed solution could process up to 50 MEPS.

The paper \cite{czech2016evaluating} presents eight filtering algorithms, four of which are used to remove background activity.
Three of them calculate the difference in timestamps between the pixel being processed and the last event in its surroundings.
The fourth method requires that at least two other events belong to the surrounding of the processed event within a given time window.

The paper \cite{barrios2018less} presents the \replace{LDSI (Less Data Same Information)}{Less Data Same Information (LDSI)} algorithm.
Its aim is to reduce the amount of data needed to be processed without removing relevant information.
The algorithm uses spiking cells inspired by the action of biological neurons to process data in four layers.
The authors were able to reduce the amount of data transmitted by 30\% while maintaining the same similarity index as\new{ the} unfiltered data.
Additionally, an architecture for the FPGA chip that implements the LDSI filter was proposed, but the maximum throughput was not specified.

The paper \cite{khodamoradi2018n} presents a spatiotemporal filter with O(N) memory complexity. The proposed method uses two memory cells for each row and column instead of one cell for each pixel, resulting in a significant reduction in memory usage. \replace{Additionally}{Furthermore}, the paper presents an FPGA implementation using high-level synthesis (HLS) tools for \replace{HDL (Hardware Description Language) code generation}{the generation of Hardware Description Language (HDL) code}, achieving a throughput of \SI{3}{MEPS} for resolutions ranging from \(128 \times 128\) to \(1280 \times 1280\).

An implementation of a filtering algorithm using a two-layer spiking neural network in the \textit{IBM TrueNorth Neurosynaptic System} neuromorphic processor is presented in \cite{padala2018noise}.
The first layer introduces a refractive period, while the second one checks whether other events have been generated in the vicinity of the event being processed.
An improvement in SNR (signal-to-noise ratio) of approximately \SI{10}{\dB} was achieved over the Nearest-Neighbour method.
The algorithm has been tested on \replace{ATIS sensor}{Asynchronous Time-based Image Sensor (ATIS)} data with a resolution of \(304 \times 240\). The throughput achieved is not specified.

A method of filtering and compression of events stream based on two switched time windows was presented in \cite{bisulco2020near}.
Events are fed into one window, and the data from the other window is processed and cleaned. In this way, an image representing the sensor data is created. Then, the operation of the windows is reversed.
The timestamps of individual events are removed and are not passed to the filter output.
The algorithm has been tested on a \(480 \times 320\) sensor and implemented on an FPGA chip. The maximum frequency of the designed architecture was \SI{51.18}{\MHz}.

In the first part of the paper \mbox{\cite{ojeda2020device}}, a point process filter (PPF) is proposed to filter out noise in the event data and generate a binary image representation.
It uses an adaptive time window and outputs a two-dimensional image in which the recorded events are marked. These are aggregated over a preset time window of \SI{3}{\ms}. In the generated image, the logical product of adjacent pixels is performed (separately in the horizontal and vertical directions).
The work uses a DVS sensor with a resolution of \(480 \times 320\). An FPGA implementation of the PPF algorithm is presented.

The authors of paper \cite{guo2020hashheat} proposed a hashing-based filtering algorithm that achieves very low memory requirements by eliminating the need \replace{for storing}{to store} 32 bit timestamps.
The algorithm was implemented on an FPGA chip with a reported operating frequency of \SI{100}{\MHz}.

The work \cite{xiao2021snn} presented a spiking neural network with an adaptive time window length, using a leaky integrate-and-fire (LIF) neuron model.
The authors claim that this method outperforms the nearest-neighbour method in terms of SNR.

In \cite{gupta2021foveal}, filtering inspired by the action of the human retina was proposed.
The retinal model used Gaussian difference filters inspired by the optic nerve fovea.
The sensor used in this work had a resolution of \(128 \times 128\).

The paper \cite{mohamed2022dba} presents an adaptive method for filtering event data.
The authors used an optical flow and a non-parametric \replace{KNN (K-nearest neighbours)}{K-nearest neighbours (KNN)} regression algorithm.
In the first stage, redundant events are removed using a technique based on a dynamic timestamp. In the second stage, the adaptive KNN algorithm is used to remove background activity noise.
The algorithm was tested on a \replace{DAVIS}{Dynamic and Active-pixel Vision Sensor (DAVIS)} with a resolution of \(240 \times 180\)\new{, which combines the functionality of DVS and active pixel sensor (APS)}.
The filter achieves a signal-to-noise ratio (SNR) of up to \SI{13.64}{\dB} in a static scene and up to \SI{6.709}{\dB} in a high dynamic scene.

A hardware architecture for event filtering was presented in\remove{ work} \cite{kowalczyk2022hardware}. The method used an IIR filter matrix for event timestamps and could process up to 385.8 MEPS.

The paper \cite{guo2022low} presents three algorithms for filtering event data.
The first algorithm calculates the distance to previous events stored in a \replace{FIFO (First-In First-Out)}{First-In First-Out (FIFO)} memory.
The second algorithm checks whether a certain number of other events were registered in the vicinity of the processed event within a given time window.
The last algorithm is based on a simple neural network that processes the age of events from the analysed environment.
Additionally, the paper suggests using the Area Under the Curve (AUC) indicator for the Receiver Operating Characteristic (ROC) to evaluate the filtering quality, regardless of the chosen discrimination threshold.
This work was extended in \mbox{\cite{Rios-Navarro_2023_CVPR}} by presenting an FPGA and \replace{ASIC}{Application-Specific Integrated Circuit (ASIC)} implementation of the MultiLayer Perceptron denoising Filter (MLPF) for a \mbox{\(346 \times 260\)} resolution sensor. The maximum event throughput for the ASIC was 25 MEPS. The average AUC for the dataset used was 0.87.

An Event Denoising Convolutional Neural Network (EDnCNN) was presented in \cite{baldwin2020event}. A \replace{dynamic active vision sensor (DAVIS)}{DAVIS} with a resolution of \(346 \times 260\) pixels was used\remove{, which combines the functionality of DVS and APS (active pixel sensors)}. EDnCNN is composed of three \(3 \times 3\) convolutional layers (using ReLU, batch normalisation, and dropout) and two fully connected layers.

A neural network based on PointNet was presented in the paper \mbox{\cite{fang2022aednet}}. The authors replaced the shared \replace{MLP}{Multilayer Perceptron} units of PointNet \replace{for}{with} 1D convolution along the time direction.
This paper also presented a \textit{DVSCLEAN} dataset for event denoising with a resolution of \mbox{\(1280 \times 720\)}.


\new{In the work \mbox{\cite{alkendi2022neuromorphic}}, a solution based on a graph neural network (GNN) transformer is presented. The GNN framework implements a message passing structure, referred to as EventConv, to reflect the spatial-temporal correlation between events while maintaining their asynchronous nature. The proposed solution could process around 19084 events per second running on a laptop with \textit{Intel core i7 7700HQ} processor and \textit{NIVIDIA GeForce GTX 1050 Ti} graphic card.}

A filtering method based on clustering using event density was presented in \cite{zhang2023neuromorphic}. \new{The} experiments were performed on a sensor with a resolution of \(346 \times 260\). The proposed solution was able to process about 2,500 events per second.

\new{The} paper \cite{xu2023denoising} proposes a method based on augmented spatiotemporal correlation. The algorithm classifies events and noise \replace{based on}{on the basis of} their spatiotemporal properties. The dynamic response characteristics of DVS in different cases are used to augment spatiotemporal correlation and construct two event filters to process different types of events.

\new{In \mbox{\cite{ding2023mlb}}, a large-scale benchmark dataset for event-based camera denoising, E-MLB, was proposed. The dataset includes 100 scenes, each with four different noise levels. A new denoising evaluation metric, the Event Structural Ratio (ESR), was also introduced. The dataset has a resolution of \mbox{\(346 \times 260\)} pixels.}

\new{The paper \mbox{\cite{duan2023neurozoom}} discusses an approach to improve the quality of neuromorphic camera data by addressing issues such as noise and low resolution. A 3D U-Net as the backbone neural architecture was used for this task. The networks are trained in a noise-to-noise fashion, where the two ends of the network are unfiltered noisy data. The proposed solution was tested on event-based visual object tracking and image reconstruction tasks, demonstrating its effectiveness in improving neuromorphic camera data for practical applications.}

\new{The} paper \cite{fang2024fast} presented an approach that focuses on\new{ the} simultaneous processing of an event stack using a neural network.
The approach includes a Temporal Window (TW) and a Soft Spatial Feature Embedding (SSFE) module to process temporal and spatial information separately. This results in a novel multiscale window-based event denoising network (WedNet) that processes the entire event window simultaneously, increasing the speed of event processing.

\new{A hardware-efficient, spatiotemporal correlation filter (HAST) was proposed to filter background activity noise near the sensor in \mbox{\cite{gopalakrishnan2024hast}}. It uses two-dimensional binary arrays along with arithmetic-free hash-based functions for storage and retrieval operations. This approach eliminates the need to use timestamps to determine the chronological order of events. HAST was implemented on an FPGA chip, achieving 18 MEPS throughput for a sensor with a resolution of \mbox{\(346 \times 260\)} pixels.}

\new{The paper \mbox{\cite{zhang2024neuromorphic}} presents a unified algorithm for neuromorphic imaging that jointly tackles image deblurring and event denoising. Since informative events are triggered at the edges of moving objects, the image gradients provide a reference for filtering out noisy events. This means that pixels with stronger intensity variations in the image are more likely to generate valid events.}

The literature review confirms the \replace{significance}{importance} and prevalence of event data filtering. Various algorithms are continually being proposed to eliminate noise.
However, these studies often neglect to address the throughput and speed of the proposed solutions.
Several papers have addressed the issue of event stream filtering using FPGA computational resources. On the other hand, most of these architectures were designed for sensors with a resolution much smaller than \(1280 \times 720\).
These considerations are significant because modern event cameras provide better and better event throughput and resolution. It is probable that these parameters will continue to improve in the future, necessitating even higher processing speed.
The use of FPGAs in the field of neuromorphic vision is promising and\new{ is} increasingly popular due to the possibility of pipeline processing of the event stream.

\section{The proposed filtering methods}
\label{sec:algorithms}
An assumed feature of the designed algorithms was low memory usage for efficient implementation in embedded circuits or FPGA reconfigurable resources. Therefore, storing information separately for each pixel (e.g. the last timestamp as in the nearest-neighbour method) is not possible. Such an approach for a sensor with a resolution of \(1280 \times 720\) would require almost 3.7 MB of memory, and for a resolution of \(640 \times 480\) more than 1.2 MB, assuming 4 bytes per pixel. In contrast, the FPGA chip used in the study has around 1.38 MB of internal BlockRAM memory. \replace{Moreover, the use of external RAM could considerably increase both latency and power consumption.}{In addition, using external RAM for data filtering could adversely affect the results obtained. This is because reading and writing data to/from external RAM takes much longer than with internal BlockRAM, and its access time is not deterministic. The latency and throughput of the designed architecture would therefore increase significantly. The use of external memory would also increase the power consumption of such a solution. It is also worth noting that this would require a memory controller, which would further increase the complexity of the architecture and its latency.}

Noise in event data is typically identified as single spikes that are not associated with a change in brightness in the observed scene. To filter it out, an algorithm should remove isolated events with very little or no other data in the neighbourhood (both in time and space).
Conversely, for motion-related events, a certain locality can usually be assumed. This means that for such events, there should be other events in the spatial-temporal neighbourhood that also respond to the same movement or change in brightness.
Therefore, it is important to analyse the neighbourhood of each event and check whether the sensor has recorded data there recently.

To \replace{decrease}{reduce} storage requirements, the entire sensor area was divided into separate subareas of equal size. Using subareas with dimensions of \(16 \times 16\) pixels,\new{ the} memory requirements can be reduced by up to 256 times, assuming 4 bytes per subarea.

\remove{The proposed algorithms store a calculated time interval between events for each area, in addition to the timestamp. This indirectly indicates the frequency of events in a given area of the sensor matrix. The higher the event frequency for a given subarea, the more weight it had in further calculations. 
Although storing this additional data increase memory usage, it reduce edge effects for events located at the edges of subareas.
The features of an area are updated for each event that is classified within it based on its location.
The updating process follows \mbox{\eqref{eq:timestamp}} and \mbox{\eqref{eq:interval}}.
This means that an infinite impulse response (IIR) filter is applied to the timestamps of the events that belong to the area and to the intervals between consecutive events.
Furthermore, a global update was implemented for inactive areas, defined as areas without any events within a given time interval. This update was intended to mitigate the impact of event deletion in long-neglected areas and to gradually decrease the interval for areas without events.}
\new{The proposed algorithms filter the timestamps for each area. The timestamp update process follows \mbox{\eqref{eq:timestamp}}.}

\begin{equation}
    \text{Ts}_{n+1} = \text{Ts}_n (1 - u) + \text{Ts}_e u
\label{eq:timestamp}
\end{equation}
where \(\text{Ts}_{n}\) represents the current timestamp of the area, \(\text{Ts}_{n+1}\) represents the new timestamp of the area, \(u\) represents the update factor, and \(\text{Ts}_e\) represents the timestamp of the processed event.

\new{In addition, an interval between events for each area is estimated. This indirectly indicates the frequency of events in a given area of the sensor matrix. The higher the event frequency for a given subarea, the more weight it has in further calculations. The interval update process follows \mbox{\eqref{eq:interval}}.}

\begin{equation}
    I_{n+1} = I_n (1 - u) + (\text{Ts}_e - \text{Ts}_n) u
\label{eq:interval}
\end{equation}
where \(I_n\) represents the current interval of the area, while \(I_{n+1}\) represents the new interval of the area.

\begin{figure}
    \centering
    \includegraphics[width=\linewidth]{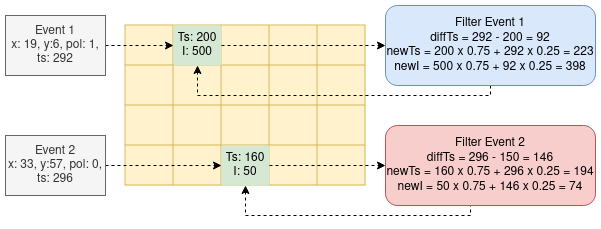}
    \caption{
    The example update of subarea parameters for two events\replace{. Each event is}{, each described with 4 values: x, y, polarisation (pol) and timestamp (ts). First, they are} assigned to an area based on pixel coordinates. The area features are then read out, and new features are calculated based on these and the timestamp of the processed event. Finally, the new features are stored.}
    \label{fig:AlgorytmFilter}
\end{figure}

\new{Although storing these additional data increases memory usage, it reduces edge effects for events located at the edges of subareas.
The features of an area are updated for each event that is classified within it based on its location.
This means that an infinite impulse response (IIR) filter is applied to the timestamps of the events that belong to the area and to the intervals between consecutive events.
Furthermore, a global update was implemented for inactive areas, defined as areas without any events within a given time interval. This update was intended to mitigate the impact of event deletion in long-neglected areas and to gradually decrease the interval for areas without events.}
Figure \ref{fig:AlgorytmFilter} shows an example of updating subareas for two events, with an update factor of 0.25.

\begin{figure}
    \centering
    \subfloat[][\label{sfig:Borders}]{\includegraphics[width=2in]{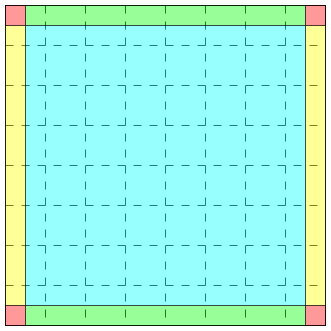}}
    \hspace{.5in}
    \subfloat[][\label{sfig:Interpolation}]{\includegraphics[width=2in]{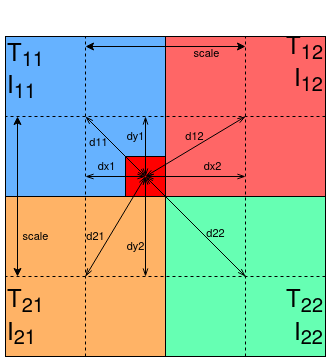}}
    \caption{
    In \ref{sfig:Borders}, the matrix division is shown as a function of the number of neighbouring regions for the event to be processed. In \ref{sfig:Interpolation} an example situation for the filtering algorithm is shown. The red pixel indicates the recorded event. Four adjacent areas also shown with their parameters and calculated quantities.
    }
    \label{fig:BordersInterpolation}
\end{figure}

Each incoming event is evaluated to determine \replace{if}{whether} it is\new{ a} noise based on the features of the surrounding areas. The analysis of the event depends on the position of the reported pixel, with 1, 2 or 4 neighbouring areas being examined, as shown in Figure \ref{sfig:Borders}. For red pixels, only one area is analysed, for yellow pixels, two areas are analysed, and for blue pixels, \replace{4}{four} neighbouring areas are analysed. Figure \ref{sfig:Interpolation} provides a more detailed illustration of the latter situation. The red pixel indicates the recorded event, with the four adjacent areas also visible. The parameters and calculated quantities of these areas are marked, with \(T\) representing the timestamps and \(I\) representing the estimated time between events. \(\text{Scale}\) refers to the size of the matrix-divided areas, while \(d\), \(\text{dx}\), and \(\text{dy}\) denote the distances from the processed event to the centres of the neighbouring areas.

Based on the research described in the paper \cite{Kowalczyk_2023_CVPR}, two of the methods proposed there were the most effective: Bilinear interpolation with frequency weights (BIF) and Distance-based interpolation with frequency weights (DIF).
The former consisted of a bilinear interpolation of timestamps, taking into account the frequency with which events occurred for neighbouring events. This is described by \eqref{eq::bilinear_weights}.

\begin{equation}
\begin{split}
    T_1 &= \frac{T_{11} I_{12} \text{dx}_2 + T_{12} I_{11} \text{dx}_1}{I_{12} \text{dx}_2 + I_{11} \text{dx}_1}\\
    T_2 &= \frac{T_{21} I_{22} \text{dx}_2 + T_{22} I_{21} \text{dx}_1}{I_{22} \text{dx}_2 + I_{21} \text{dx}_1}\\
    T &= \frac{T_1 I_{21} I_{22} \text{dy}_2 + T_2 I_{11} I_{12} \text{dy}_1}{I_{21} I_{22} \text{dy}_2 + I_{11} I_{12} \text{dy}_1}
\end{split}
\label{eq::bilinear_weights}
\end{equation}

The formula designations are consistent with those in Figure \ref{sfig:Interpolation}.

The second method also \replace{utilises}{uses} the frequency of neighbouring areas, but instead of interpolating horizontally and vertically, it calculates the distance to the centres of the areas using the L2 norm, as described in \eqref{eq::distance_C} and \eqref{eq::distance_T}. The timestamps of neighbouring areas are weighted proportionally to the frequency of the events and inversely proportional to their distance from the event. Proportionality to frequency also implies inverse proportionality to the interval between events, which is directly estimated.

\begin{equation}
\begin{split}
    C_{11} &= \frac{1}{I_{11} d_{11}}\quad
    C_{12} = \frac{1}{I_{12} d_{12}}\\
    C_{21} &= \frac{1}{I_{21} d_{21}}\quad
    C_{22} = \frac{1}{I_{22} d_{22}}
\end{split}
\label{eq::distance_C}
\end{equation}

\begin{equation}
    T = \frac{T_{11} C_{11} + T_{12} C_{12} + T_{21} C_{21} + T_{22} C_{22}}{C_{11} + C_{12} + C_{21} + C_{22}}
\label{eq::distance_T}
\end{equation}

Equation \eqref{eq::comp} compares the difference between the timestamp of the processed event and the result of algorithm \(T\) with the selected threshold, known as the filter length. If the difference is greater, the event is passed on; otherwise, it is rejected.

\begin{equation}
    R = F_L > \text{Ts} - T
\label{eq::comp}
\end{equation}
where \(R\) is the binary value of the filter result, \(\text{Ts}\) is the timestamp of the processed event and \(F_L\) is the length of the filter.

Both algorithms were modified to incorporate the subtraction from the formula \mbox{\eqref{eq::comp}} into the formulae \mbox{\eqref{eq::bilinear_weights}} and \mbox{\eqref{eq::distance_T}}. This resulted in the formulas \mbox{\eqref{eq::time_BIF}} for the BIF algorithm and \mbox{\eqref{eq::time_DIF}} for the DIF algorithm.
This means that the timestamps of neighbouring regions can be subtracted from the timestamp of the event and\remove{ further} processed in the same way without changing the result.

\begin{equation}
\begin{split}
    \Delta T_2 &= \frac{\Delta T_{21} I_{22} \text{dx}_2 + \Delta T_{22} I_{21} \text{dx}_1}{I_{22} \text{dx}_2 + I_{21} \text{dx}_1}\\
    \Delta T_1 &= \frac{\Delta T_{11} I_{12} \text{dx}_2 + \Delta T_{12} I_{11} \text{dx}_1}{I_{12} \text{dx}_2 + I_{11} \text{dx}_1}\\
    \text{Ts} - T &= \frac{\Delta T_1 I_{21} I_{22} \text{dy}_2 + \Delta T_2 I_{11} I_{12} \text{dy}_1}{I_{21} I_{22} \text{dy}_2 + I_{11} I_{12} \text{dy}_1}
\end{split}
\label{eq::time_BIF}
\end{equation}

\begin{equation}
    \text{Ts} - T = \frac{\Delta T_{11} C_{11} + \Delta T_{12} C_{12} + \Delta T_{21} C_{21} + \Delta T_{22} C_{22}}{C_{11} + C_{12} + C_{21} + C_{22}}
\label{eq::time_DIF}
\end{equation}

\begin{equation}
    \begin{split}
        \Delta T_{11} &= \text{Ts} - T_{11}\qquad
        \Delta T_{12} =  \text{Ts} - T_{12}\\
        \Delta T_{21} &= \text{Ts} - T_{21}\qquad
        \Delta T_{22} =  \text{Ts} - T_{22}
    \end{split}
\label{eq::deltaT}
\end{equation}

It may appear to be an illogical approach to transform a single subtraction in\remove{ the} \mbox{\eqref{eq::comp}} into four subtractions in \mbox{\eqref{eq::deltaT}}. However, it is essential to remember that the objective of this algorithm, in addition to efficient filtering, is to optimally use the computational resources of the FPGA device. This modification enables the range of variables on which operations are performed to be reduced, which in fixed-point format translates into a smaller number of bits, thus reducing resource usage and increasing the maximum frequency of operation.

For both methods, it was decided to perform transformations of \mbox{\eqref{eq::bilinear_weights}}, \mbox{\eqref{eq::distance_T}} and \mbox{\eqref{eq::comp}} to eliminate divisions. This decision was made because divisions have significantly higher latency and require more computational resources than multiplications and addition operations.

For the BIF, additional designations of \mbox{\eqref{eq:BIF_SC}} were adopted.

\begin{equation}
\begin{aligned}
    B_{11}  &= I_{12} \text{dx}_2 \\
    B_{12}  &= I_{11} \text{dx}_1 \\
    S_{top} &= B_{11} + B_{12} \\
    B_{top} &= I_{21} I_{22} \text{dy}_2
\end{aligned}
\qquad
\begin{aligned}
B_{21}  &= I_{22} \text{dx}_2\\
B_{22}  &= I_{21} \text{dx}_1\\
S_{bot} &= B_{21} + B_{22}\\
B_{bot} &= I_{11} I_{12} \text{dy}_1
\end{aligned}
\label{eq:BIF_SC}
\end{equation}
\begin{equation*}
S_{all} = B_{top} + B_{bot}
\end{equation*}

Then by multiplying both sides of the final comparison by the introduced variables \mbox{\(S_{all}\)}, \mbox{\(S_{top}\)}, \mbox{\(S_{bot}\)}, it can be written as \mbox{\eqref{eq:BIF_multiply}}.

\begin{equation}
\begin{split}
    F_c = &F_L S_{all} S_{top} S_{bot}\\
    \Delta T_c = &(\Delta T_{11} B_{11} + \Delta T_{12} B_{12}) S_{bot} B_{top} + \\
         &(\Delta T_{21} B_{21} + \Delta T_{22} B_{22}) S_{top} B_{bot}\\
    R =  &F_c > \Delta T_c
\end{split}
\label{eq:BIF_multiply}
\end{equation}

For DIF, additional designations of \mbox{\eqref{eq:DIF_SC}} have been adopted.

\begin{equation}
\begin{split}
    D_{11}  &= I_{12} d_{12} I_{21} d_{21} I_{22} d_{22}\\
    D_{12}  &= I_{11} d_{11} I_{21} d_{21} I_{22} d_{22}\\
    D_{21}  &= I_{11} d_{11} I_{12} d_{12} I_{22} d_{22}\\
    D_{22}  &= I_{11} d_{11} I_{12} d_{12} I_{21} d_{21}\\
    D_{sum} &= D_{11} + D_{12} + D_{21} + D_{22}
\end{split}
\label{eq:DIF_SC}
\end{equation}

Then, by multiplying both sides of the final comparison by the introduced variable \mbox{\(D_{sum}\)}, it can be written as \mbox{\eqref{eq:DIF_multiply}}.

\begin{equation}
\begin{split}
    F_c = &F_L D_{sum}\\
    \Delta T_c = &\Delta T_{11} D_{11} + \Delta T_{12} D_{12} + \\
         &\Delta T_{21} D_{21} + \Delta T_{22} D_{22}\\
    R =  &F_c > \Delta T_c
\end{split}
\label{eq:DIF_multiply}
\end{equation}

Upon analysing the above transformations, it can be calculated that the BIF method requires 19 multiplications and 7 additions. In contrast, the DIF method requires 15 multiplications and 7 additions, assuming\new{ that} additional variables are added to \mbox{\eqref{eq:DIF_SC}} to avoid redundant calculations.

As these algorithms are the basis of the proposed hardware architecture with the highest possible throughput, it is also worthwhile to analyse their suitability for fixed-point calculations. In this regard, it is necessary to take into account the possible ranges of variables that may be used by the algorithm.

Both versions were implemented and the range of all variables in these methods was compared for\new{ the} datasets described in Section \mbox{\ref{sec:datasets}}. In the BIF method, the maximum value of variable \mbox{\(F_c\)} was \mbox{\(1.01 \cdot 10^{37}\)}, which translates to 123 bits in fixed-point format. In the DIF method, the maximum value of variable \mbox{\(F_c\)} was \mbox{\(4.37 \cdot 10^{30}\)}, which translates to 102 bits in fixed-point format.

Analysing both the number of computations required and the range of values of the variables used for the BIF and DIF methods considered, it can be concluded that the DIF method is more suitable for designing a computational architecture for FPGA resources.

\section{Datasets}
\label{sec:datasets}
Designing and testing event stream filtering algorithms can be challenging due to the difficulty \replace{in}{of} evaluating their effectiveness. This is because it is not always easy to determine whether an event is a noise or whether it is caused by movement or changes in brightness in the observed scene.
The methodology presented in\remove{ the paper} \cite{guo2022low} was inspired to solve this problem. To investigate the\remove{ filter} effectiveness\new{ of the filter} under different conditions and for different noise intensities, several elements are required. 
New datasets were prepared due to the unavailability of a dataset with a resolution of \(1280 \times 720\). This also enabled the preparation of sequences with varying levels of movement in the observed scene and\new{ of} the sensor itself.

The initial requirement is a dataset that is free from noise. Such data cannot be captured with a physical camera, but can be synthetically generated from a frame camera recording using simulators such as V2E \cite{hu2021v2e}. For this study, four \replace{data sets}{sequences} with a resolution of \(1280 \times 720\) and a length of \SI{50}{\s} were generated. Events occurring between tenth and twentieth seconds were then selected to reduce the amount of data for further processing. The temporal resolution was set to \SI{100}{\us}. The values were selected to ensure compatibility with modern event sensors. However, a temporal resolution of \SI{1}{\us} was not feasible due to excessive computational complexity.

\textit{101}:
The first recording was generated for a camera \replace{moved}{moving} in a lecture hall, with no other movement except that of the sensor.\new{ The original frame rate was equal to 50 frames per second.} The mean and median sparsity of the\remove{ data} generated\new{ data} are \(89.9\)\% and \(90.2\)\%, respectively. Sparsity refers to the proportion of pixels that do not contain any events within \SI{20}{\ms} time windows.

\textit{Library}:
The second video presents data recorded in an urban environment with low traffic.\new{ The original frame rate was equal to 25 frames per second.} The camera and a few people visible in the footage are in motion. The majority of the scene is occupied by buildings, parked cars, and trees. The\remove{ data} generated\new{ data} has a mean sparsity of \(87.3\)\% and a median sparsity of \(88.1\)\%.

\textit{Model}:
The third recording shows a camera moving along a \replace{mock-up}{simulated} street with traffic lanes.\new{ The original frame rate was equal to 30 frames per second.} The source of motion is solely the moving sensor. \replace{Additionally}{In addition}, there is noticeable camera jitter in this recording. The generated data has a mean sparsity of \(78.6\)\% and a median sparsity of \(79.0\)\%.

\textit{RIS}:
The final footage is also captured in an urban environment. However, in addition to the camera's movement, there are several moving objects in the scene, including cars and people.\new{ The original frame rate was equal to 25 frames per second.} This recording is not publicly available due to the privacy of the individuals visible in the footage. The generated data has a mean sparsity of \(86.0\)\% and a median sparsity of \(87.4\)\%.

\begin{figure}
    \centering
    \subfloat[][\label{sfig:101}]{\includegraphics[width=2in]{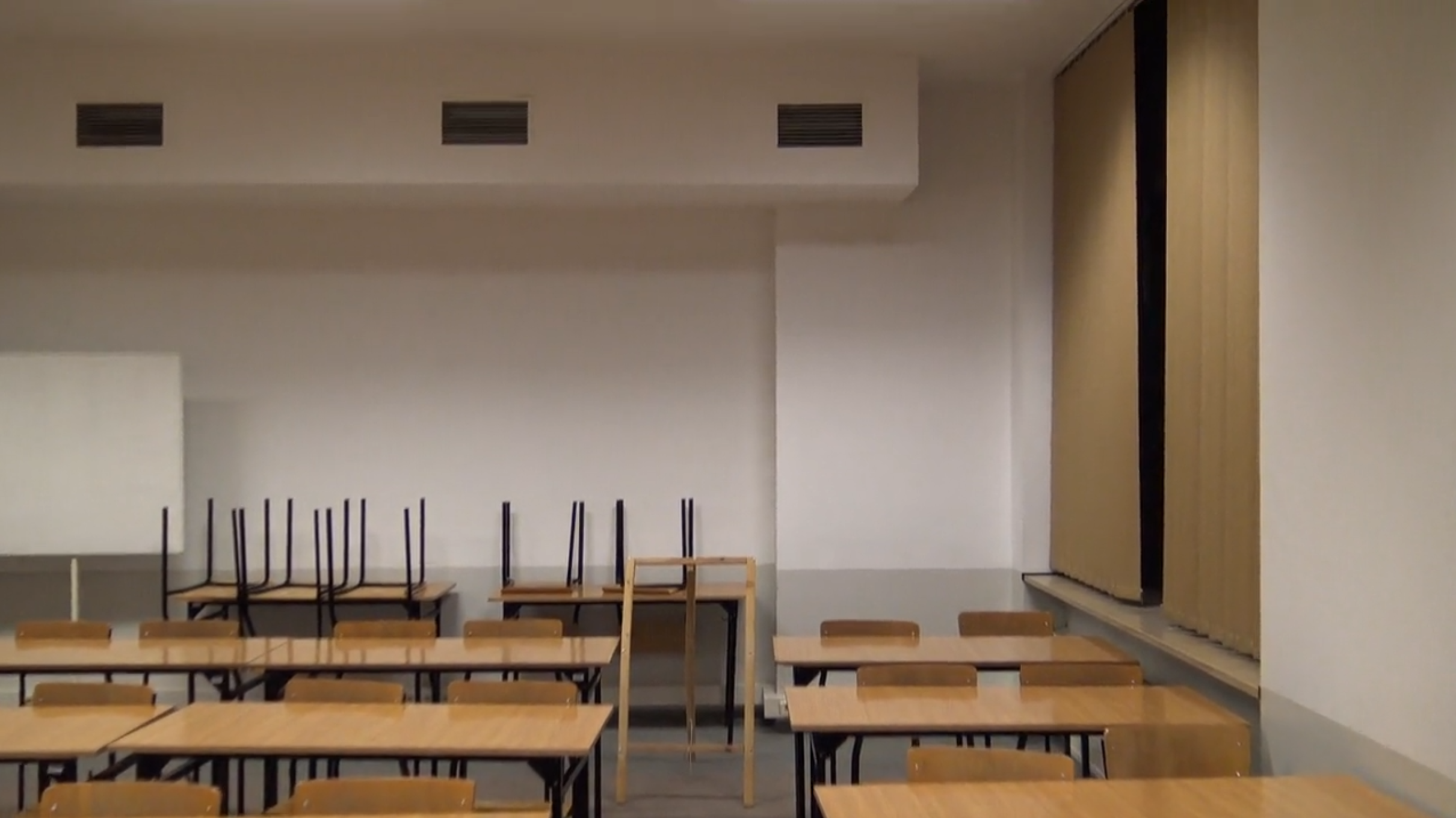}}
    \hspace{.3in}
    \subfloat[][\label{sfig:Biblioteka}]{\includegraphics[width=2in]{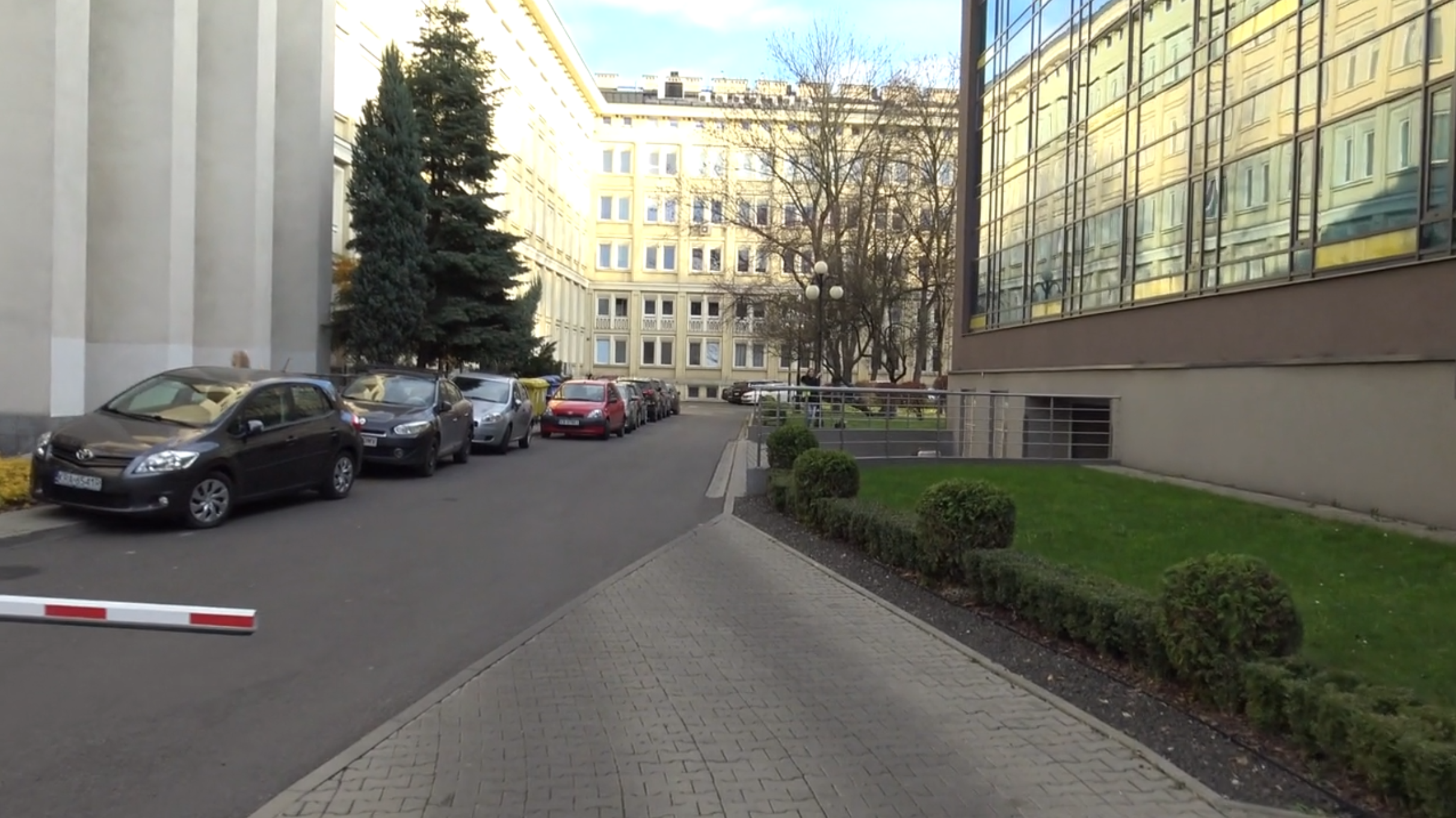}}

    \subfloat[][\label{sfig:Makieta}]{\includegraphics[width=2in]{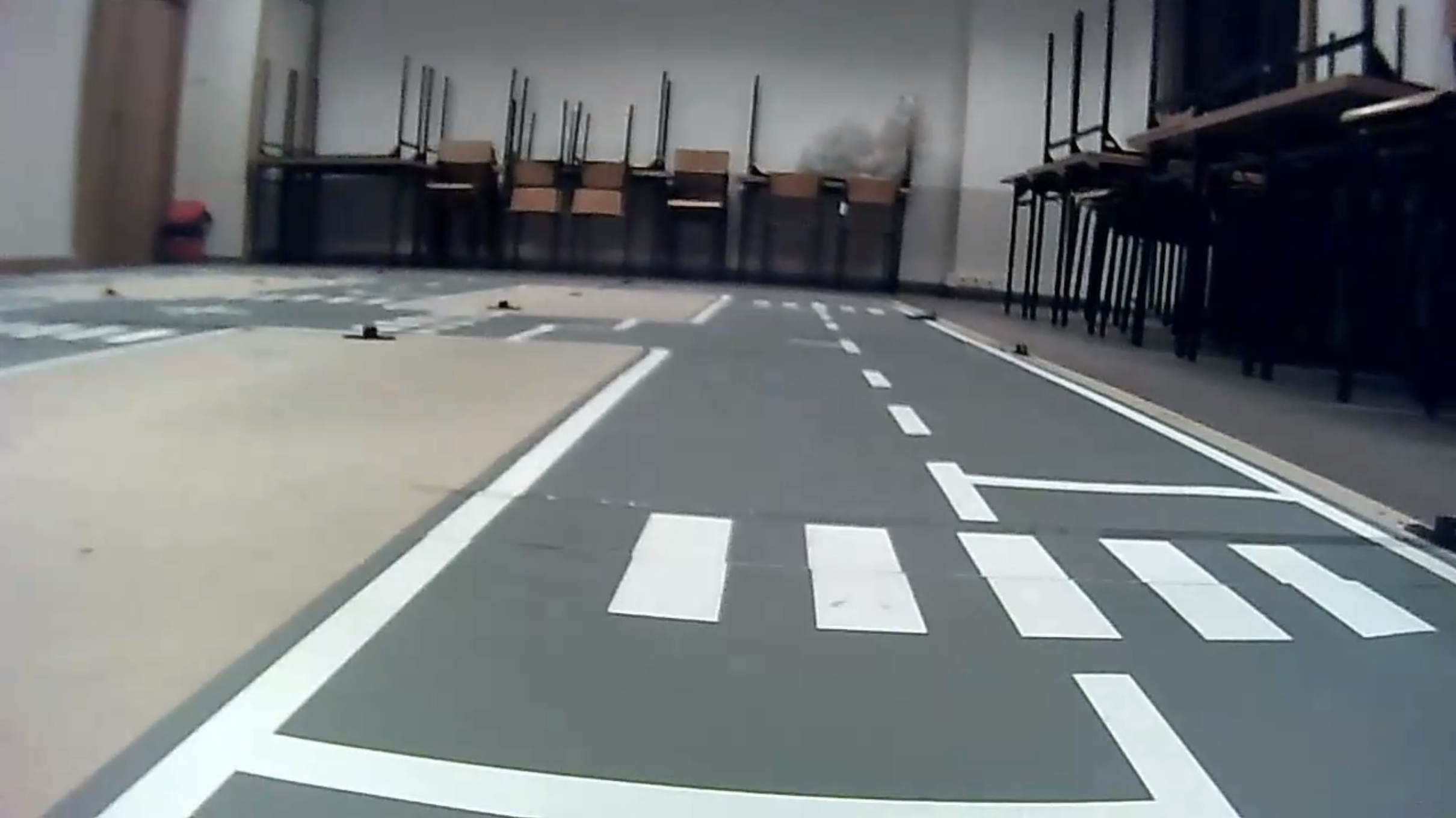}}
    \hspace{.3in}
    \subfloat[][\label{sfig:RIS}]{\includegraphics[width=2in]{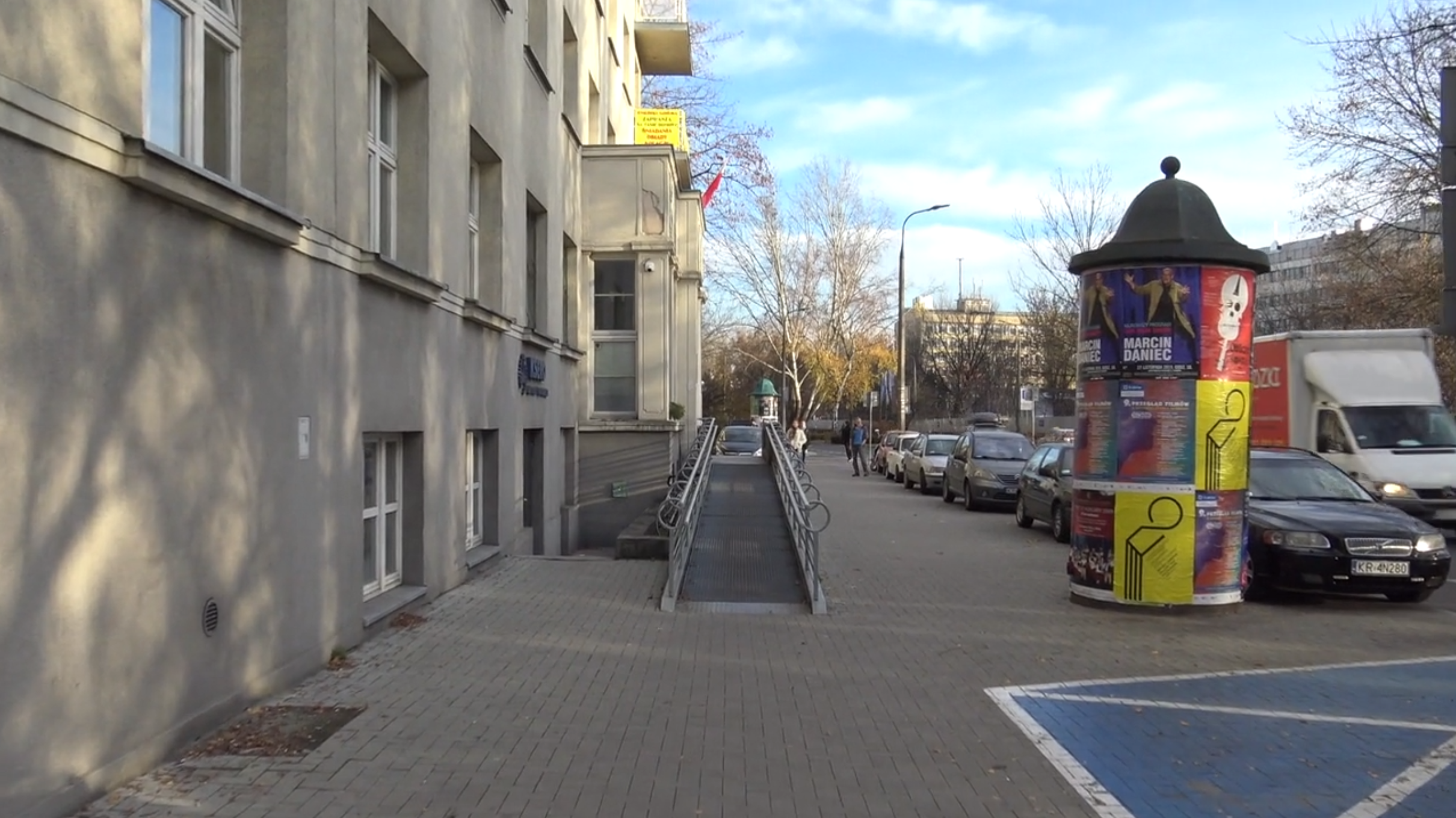}}
    \caption{
    Example frames from input recordings to the \textit{V2E} event camera simulator. \ref{sfig:101} shows a recording from inside a building, \ref{sfig:Biblioteka} a recording from outside and a small amount of motion, \ref{sfig:Makieta} a recording from a mock-up street with a lot of vibration, \ref{sfig:RIS} a recording from outside with a lot of motion.
    }
    \label{fig:V2EDatasetsInput}
\end{figure}

Example screenshots of the input recordings are shown in Figure \ref{fig:V2EDatasetsInput}.

\begin{figure}
    \centering
    \subfloat[][\label{sfig:101O}]{\includegraphics[width=2in]{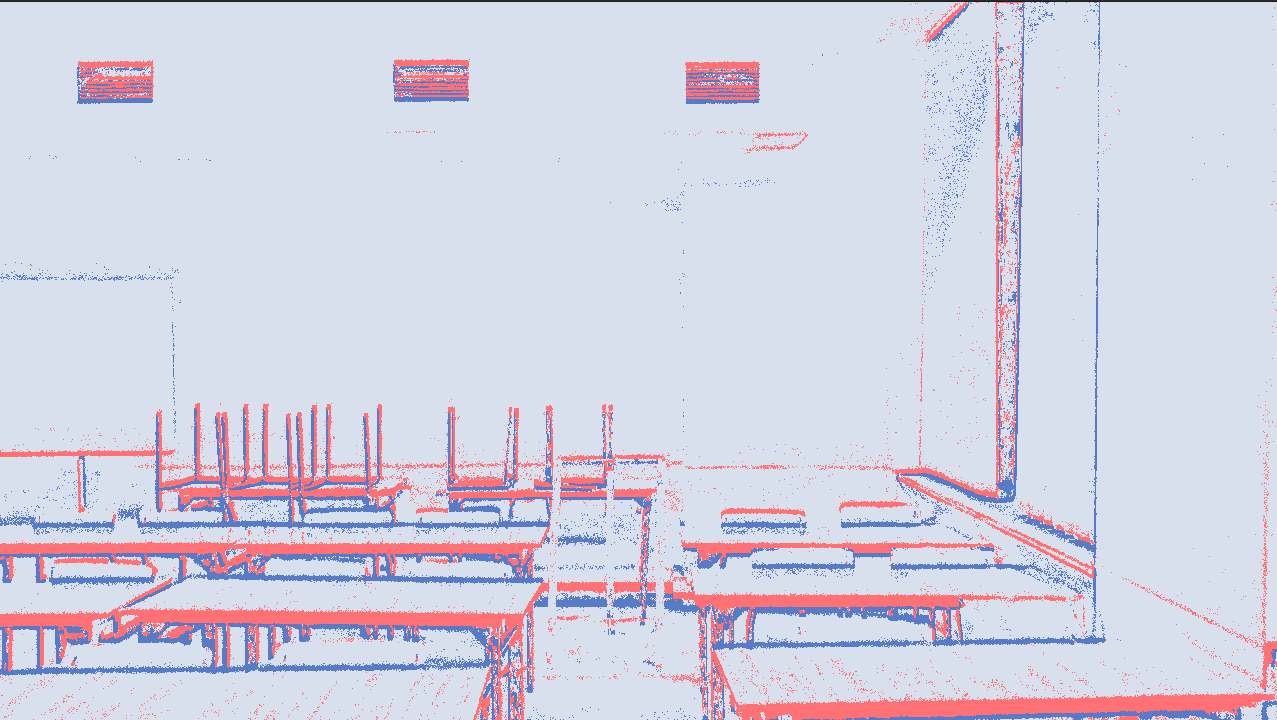}}
    \hspace{.3in}
    \subfloat[][\label{sfig:BibliotekaO}]{\includegraphics[width=2in]{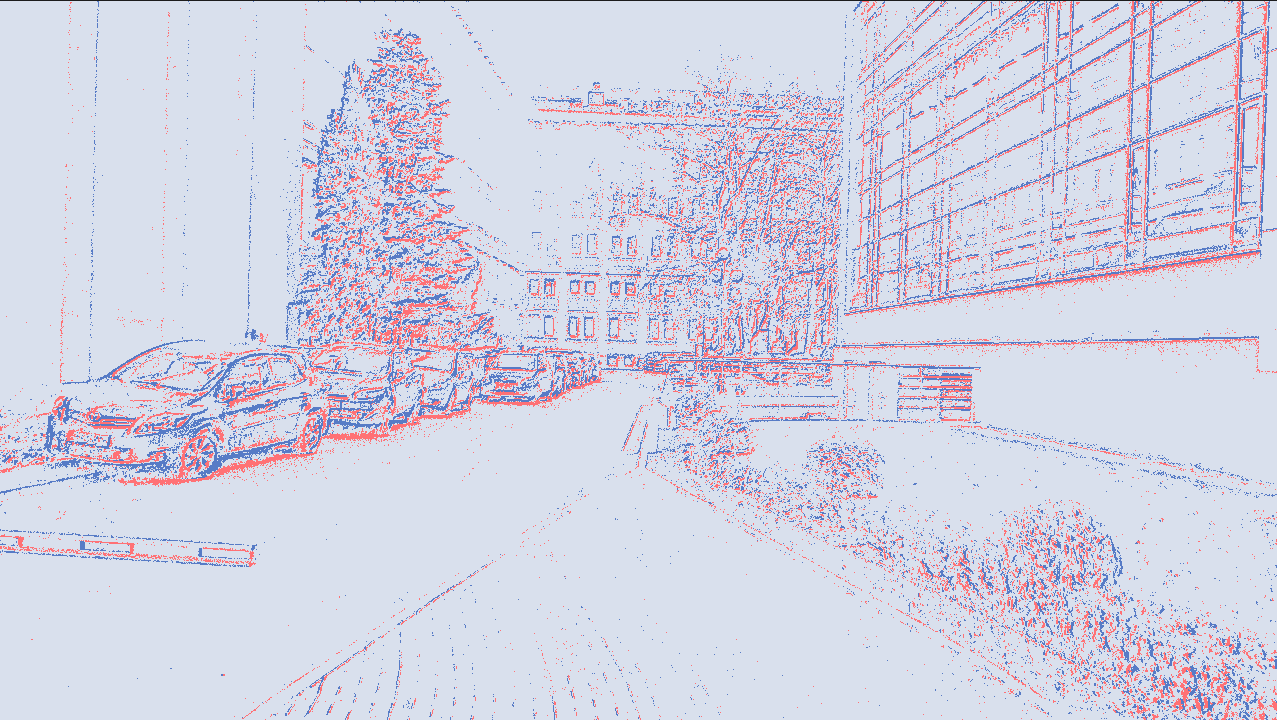}}

    \subfloat[][\label{sfig:MakietaO}]{\includegraphics[width=2in]{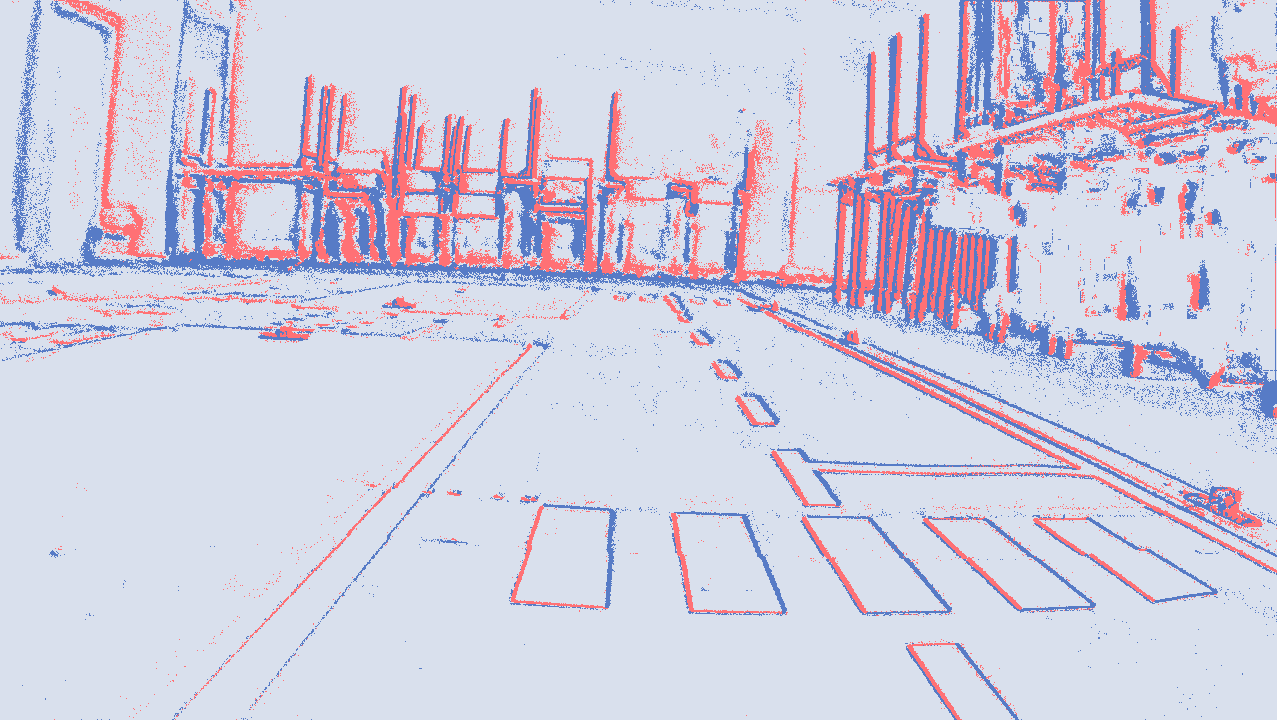}}
    \hspace{.3in}
    \subfloat[][\label{sfig:RISO}]{\includegraphics[width=2in]{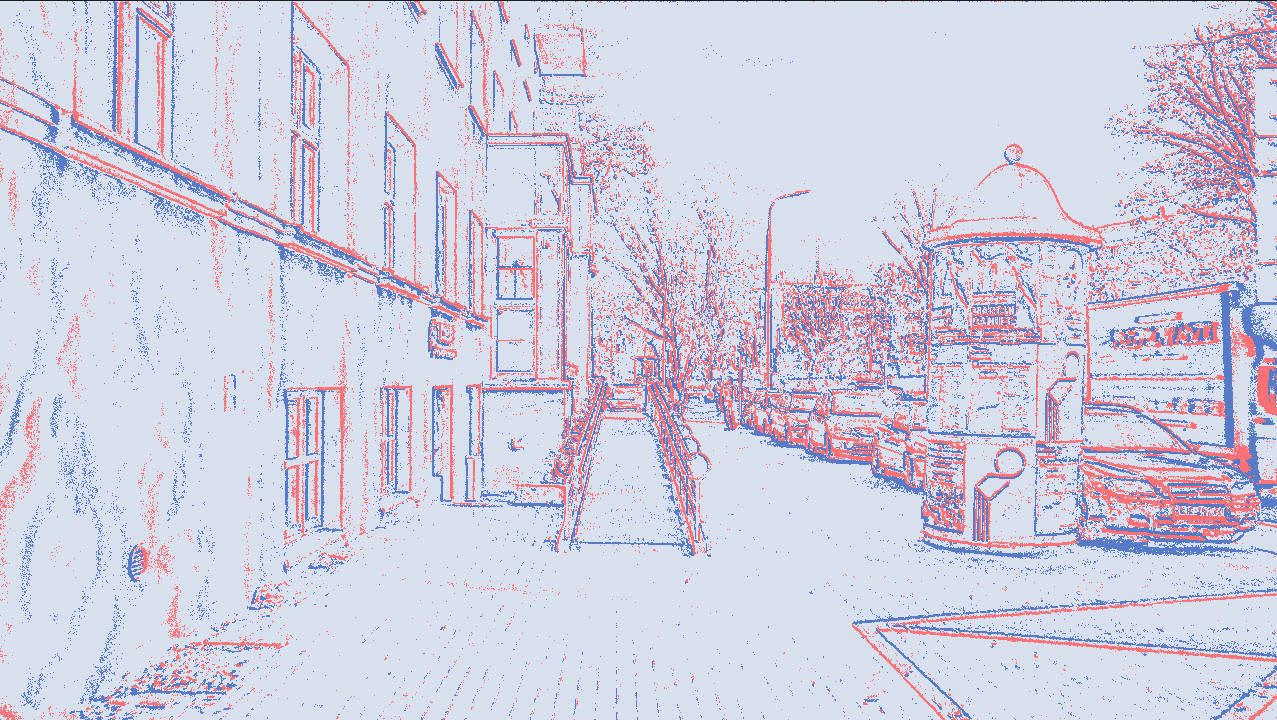}}
    \caption{
    Event frames created from generated data with an accumulation time of \SI{20}{\ms}.
    }
    \label{fig:V2EDatasetsOutput}
\end{figure}

Figure \ref{fig:V2EDatasetsOutput} shows example frames created from the generated events, corresponding to those in Figure \ref{fig:V2EDatasetsInput}.

The second component of the system is the noise generator, which is necessary to produce known noise that can be combined with ``clean'' recordings. A noise generator, based on the one proposed in \cite{guo2022low}, was implemented using the C++ language and the Metavision library developed by Prophesee.
\new{The noise was generated with a Poisson process which was approximated with a sequence of Bernoulli trials. All pixels have the same average noise rate.}
\replace{The source code for the generator}{The generator source code} has been uploaded to the GitHub repository: \url{https://github.com/vision-agh/NoiseGeneratorCPP}\footnote{The material will be published once the paper has been accepted.}. The generated events are labelled by modifying their polarity values to enable differentiation in the processed data. The polarity values of 0 and 1 are transformed to 2 and 3, \replace{correspondingly}{respectively}. The generator \replace{permits}{allows for} adjustment of the noise intensity, facilitating testing of the algorithm under varying conditions, and assessing the impact of noise on its efficacy. Noise was generated \replace{across}{at} frequencies ranging from \SI{0.01}{\Hz/px} to \SI{5}{\Hz/px}.

The event camera recordings, which are the third element, must be filtered aggressively to generate real recordings with minimal noise.
To ensure accurate evaluation of the algorithm, it is necessary to include datasets recorded with real cameras, as those generated with an event camera model may not ideally reflect reality. There is a risk that the recorded data may produce different results than the generated data.
To achieve this, we prepared several sequences using Prophesee's \textit{EVK1 - HD} event camera and applied the nearest-neighbour (NNb) algorithm to remove noise. The length of the filter was selected for each recording separately \replace{based on}{on the basis of} a subjective evaluation of the quality of the result.
\new{Each recording was processed with a set of filter lengths. The results were displayed as an event frame with \mbox{\SI{10}{\ms}} accumulation time. The authors looked at each result for an analysed sequence and selected the filter length that removed the vast majority of visible noise while preserving as many valuable events as possible.}
Both the original recordings and the filtered ones were stored in a repository (directories \textit{Recorded} and \textit{Recorded\_filtered})\footnote{The material will be published once the paper has been accepted.}.

\textit{Corn}:
In the first image, a statically mounted camera observed falling maize grains. The filter length was \SI{200}{\us}. The mean and median sparsity are \(99.8\)\% i \(100.0\)\% respectively.

\textit{Street}:
The second recording features a walk \replace{along}{on} an urban sidewalk with a filter length of \SI{200}{\us}. The sparsity values are \(92.4\)\% for the mean and \(91.5\)\% for the median.

\textit{Dancing}:
The third recording captured the dancing people using a static camera, although slight camera shake may have occurred. The filter length used was \SI{1000}{\us}. The sparsity mean and\new{ the} median were both \(97.6\)\%.

\textit{People}:
In the last one, a group of standing and moving people \replace{were}{was} recorded with a moving camera. The filter length was \SI{500}{\us}. The mean and median sparsity are \(96.1\)\% and \(96.4\)\% respectively.

\begin{figure}
    \centering
    \subfloat[][\label{sfig:Liczenie_1}]{\includegraphics[width=2in]{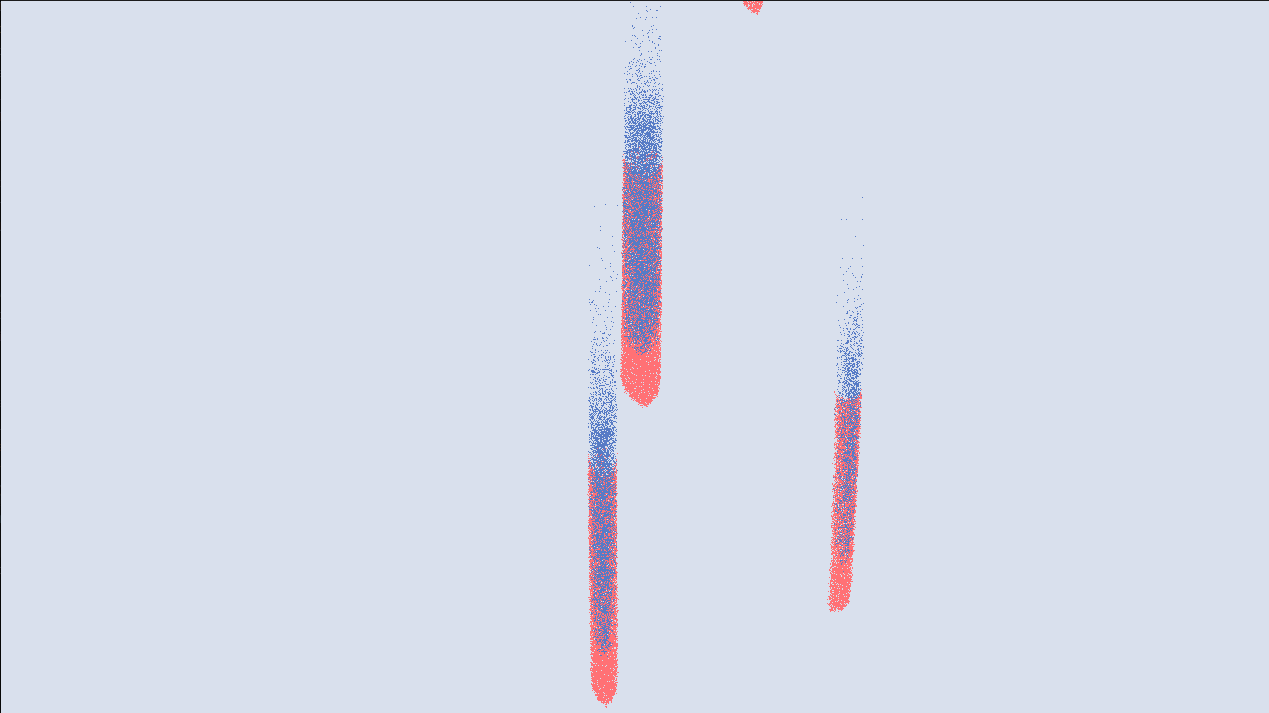}}
    \hspace{.3in}
    \subfloat[][\label{sfig:Street3}]{\includegraphics[width=2in]{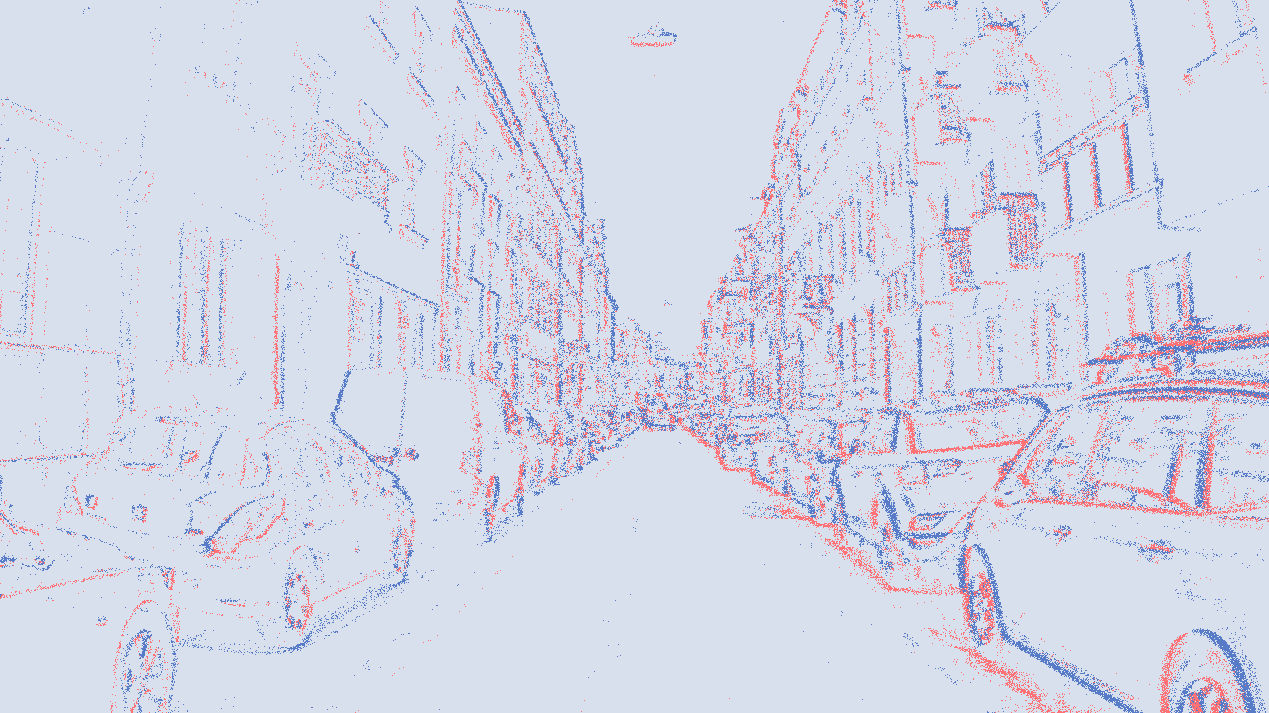}}

    \subfloat[][\label{sfig:DancingPeople3}]{\includegraphics[width=2in]{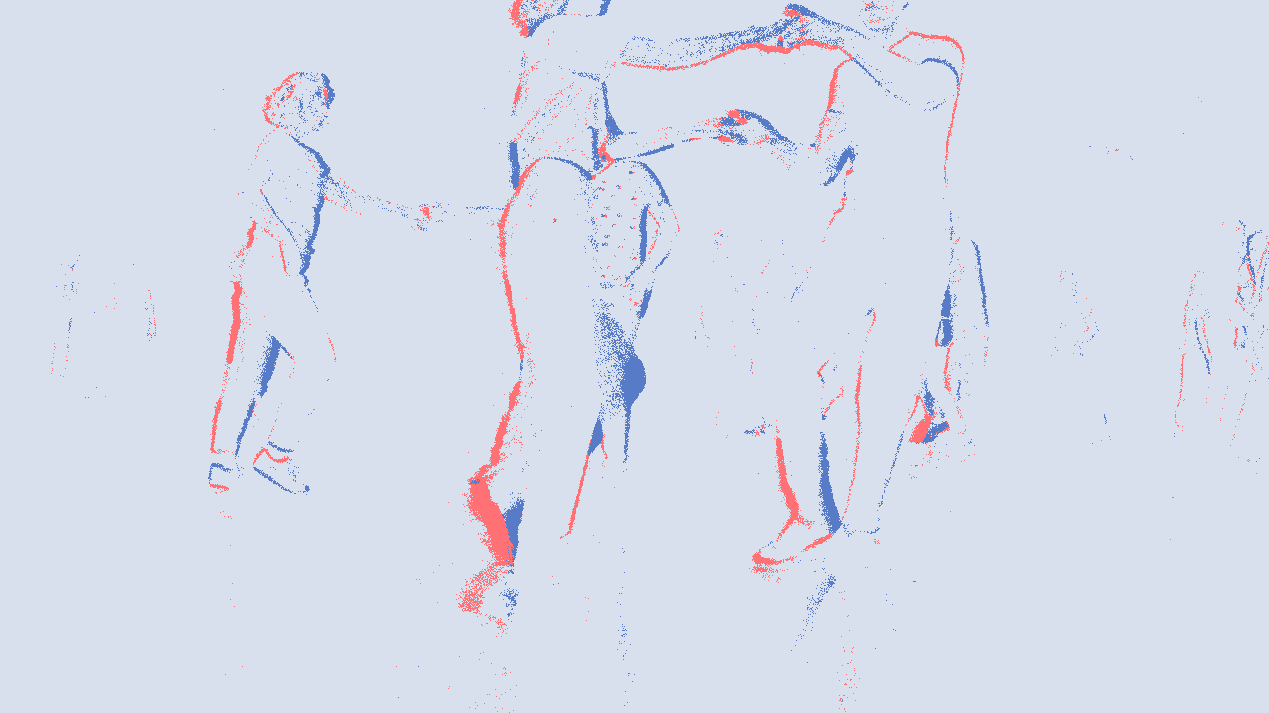}}
    \hspace{.3in}
    \subfloat[][\label{sfig:DancingCameraMoving2}]{\includegraphics[width=2in]{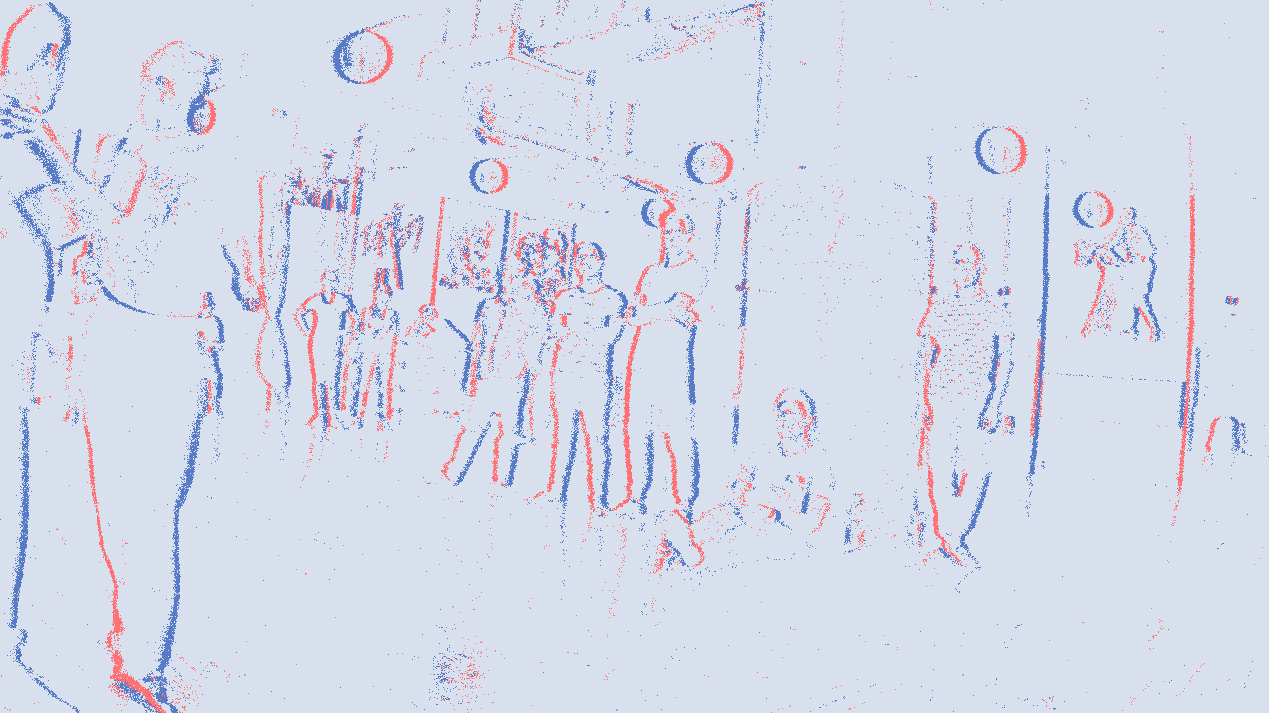}}
    \caption{
    Event frames created from recorded data with an accumulation time of \SI{20}{\ms}.
    }
    \label{fig:EVK4Dataset}
\end{figure}

Figure \ref{fig:EVK4Dataset} shows sample event frames created from the recorded sequences.

The last component comprises the recordings that contain noise from the event camera, which will be merged with the clean recordings.
Similarly to the earlier situation, it is important to also perform tests with the noise recorded by a real sensor, as the generated may not be ideal. Again, there is a risk that the recorded noise may produce different results than the generated ones.
The camera captured a static scene under varying light levels, at different operating times (immediately after turning on the sensor or after waiting a few minutes for its temperature to stabilise) and with different sensor operating parameters. Real noise was obtained with varying degrees of intensity ranging from \SI{0.0038}{\Hz/px} to \SI{1.82}{\Hz/px}. It was placed in the same repository (\textit{Recorded\_noise}).

Links to the prepared datasets were made available in the GitHub repository, https://github.com/vision-agh/FilteringDatasets\footnote{The material will be published once the paper has been accepted.}, provided there was no risk of third-party image sharing (i.e. other people in public spaces).
\replace{In comparison}{Compared} to the work \cite{Kowalczyk_2023_CVPR}, the ability to generate noise of any intensity and test sequences without noise generated using \textit{V2E} has been added.

\section{Evaluation}
\label{sec:evaluation}
The effectiveness of the algorithm was \replace{assessed}{evaluated} using the Area Under Receiver Operating Characteristic (AUROC) index, which was proposed to evaluate the effectiveness of event data filtering in the article \cite{guo2022low}. The ROC graph displays the ratio of false positives (i.e. the proportion of noise that passed through the filter output) on the horizontal axis and the ratio of true positives (i.e. the proportion of correct events that passed through the filter output) on the vertical axis for different thresholds. The AUROC value indicates the effectiveness of the filter regardless of the threshold chosen.

The filters and functions used to evaluate the effectiveness of filtering were implemented in C++ using the \textit{Pipeline} feature of the \textit{Metavision} library. The choice of these tools was based on several reasons, the most significant being that the \textit{Metavision} library is dedicated to \textit{Prophesee}'s event sensors. Furthermore, the designed solutions can be easily executed with data from other DVS sensors. This is possible because the \textit{Metavision} library allows loading data from \textit{dat} files, which have a simple binary format, or the widely used \textit{hdf5} dataset format. \replace{Additionally}{Furthermore}, the combination of the C++ language and the \textit{Pipeline} functionality, which supports multithreading, significantly enhances the efficiency of the calculations performed.
The software was executed on the \textit{Ares} supercomputer at the \textit{Cyfronet} academic computer centre due to the extensive calculations required.

\begin{figure}
    \centering
    \includegraphics[width=0.8\linewidth]{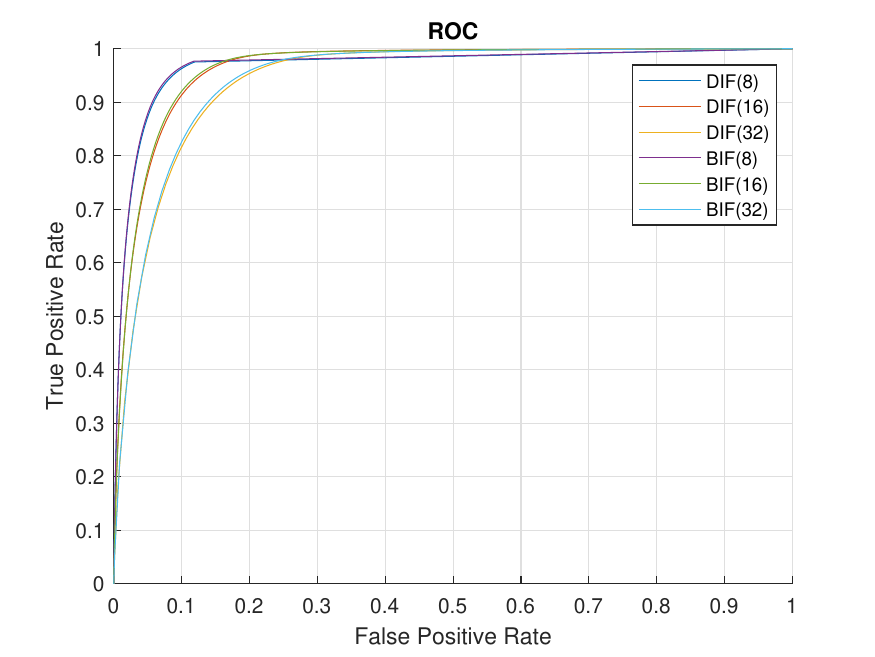}
    \caption{
    An example ROC plot that compares the performance of DIF and BIF filters for the \textit{People} dataset at a noise intensity of \SI{1}{\Hz/px}.}
    \label{fig:ROC_People}
\end{figure}

Figure \ref{fig:ROC_People} shows an example of ROC plots for the \textit{People} dataset with a noise intensity of \SI{1}{\Hz/px}. The results for\new{ the} DIF and BIF filters are compared using an update factor of 0.25, scale parameters of 8, 16, and 32, and a global update every \SI{20}{\ms}.
AUC values for these characteristics range from \(0.943\) to \(0.969\).
In this case, the BIF filter with a scale parameter of 8 was found to be the most effective.

\begin{figure}
    \centering
    \includegraphics[width=0.8\linewidth]{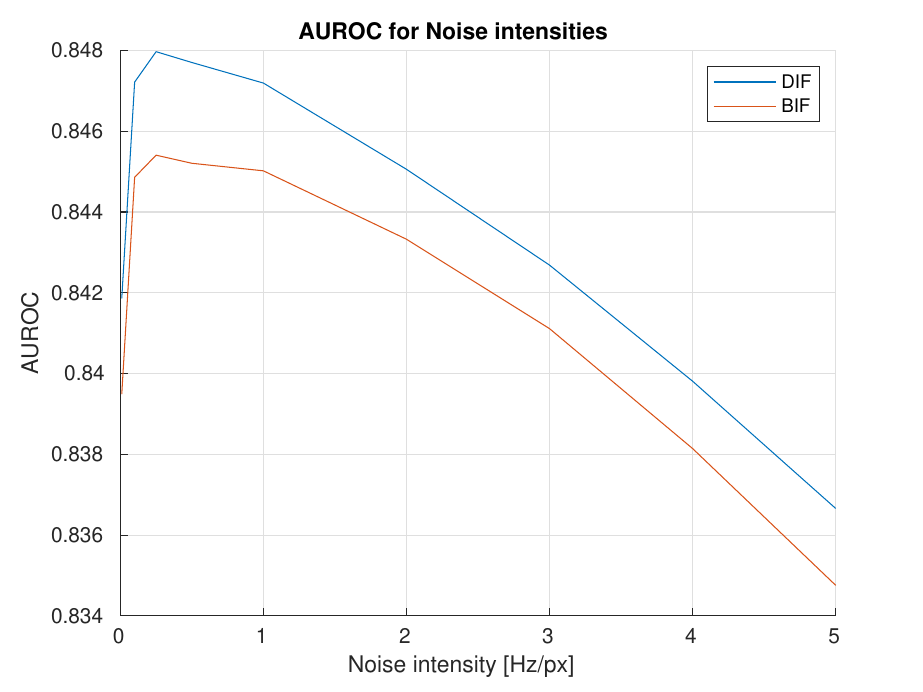}
    \caption{
    The plot displays the AUROC coefficient for various noise intensities in the \textit{Street} dataset.}
    \label{fig:AUROC_Street}
\end{figure}

The impact of noise intensity on the effectiveness of the filters was analysed. Graphs were used to illustrate the changes in the AUROC index for the datasets. Figure \ref{fig:AUROC_Street} shows an example graph for the \textit{Street} dataset.
\replace{Additionally}{In addition}, the average\new{ value of the} AUROC index was calculated for each \replace{test set}{set of tests}. Table \ref{tab:AUROC_generatedNoise} displays the calculated values for datasets with artificially generated noise, while Table \ref{tab:AUROC_recordedNoise} presents these values for datasets with recorded noise. The update factor was assumed to be 0.25, with a scale parameter of 16 and a global update occurring every \SI{20}{\ms}.

\begin{table}[!t]
    \centering
    \caption{Average AUROC values for datasets with generated noise.}
    \begin{tabular}{| c | c | c |}
        \hline 
        \textit{Dataset} & DIF   & BIF  \\	
        \hline
        \textit{101}     & 0.946 & \textbf{0.948}\\
        \hline
        \textit{Library} & \textbf{0.863} & 0.861  \\
        \hline
        \textit{Model}   & \textbf{0.886} & 0.884\\
        \hline
        \textit{RIS}     & \textbf{0.856} & 0.855   \\
        \hline
        \textit{Corn}    & \textbf{0.999} & 0.999   \\
        \hline
        \textit{Street}  & \textbf{0.844} & 0.842 \\
        \hline
        \textit{Dancing} & 0.960 & \textbf{0.961}  \\
        \hline
        \textit{People}  & 0.900 & \textbf{0.902} \\
        \hline
        Average          & \textbf{0.907} & 0.907 \\
        \hline
    \end{tabular}
    \label{tab:AUROC_generatedNoise}
\end{table}

\begin{table}[!t]
    \centering
    \caption{Average AUROC values for datasets with recorded noise.}
    \begin{tabular}{| c | c | c |}
        \hline 
        \textit{Dataset} & DIF   & BIF \\	
        \hline
        \textit{101}     & 0.942 & \textbf{0.944} \\	
        \hline
        \textit{Library} & \textbf{0.873} & 0.870   \\
        \hline
        \textit{Model}   & \textbf{0.888} & 0.887 \\
        \hline
        \textit{RIS}     & \textbf{0.855} & 0.853    \\
        \hline
        \textit{Corn}    & \textbf{0.999} & 0.999    \\
        \hline
        \textit{Street}  & \textbf{0.856} & 0.853  \\
        \hline
        \textit{Dancing} & 0.961 & \textbf{0.962}   \\
        \hline
        \textit{People}  & 0.896 & \textbf{0.898}  \\
        \hline
        Average          & \textbf{0.909} & 0.908  \\
        \hline
    \end{tabular}
    \label{tab:AUROC_recordedNoise}
\end{table}

Upon analysis of the data presented, it is evident that both proposed methods yield comparable results. It is worth noting that these methods perform better in scenarios where the camera is stationary and the observed scene is in motion.

Finally, the DIF method was selected \replace{as}{because} it has \replace{a similar efficiency as}{an efficiency similar to that of} BIF and requires less hardware resources (cf. Section \mbox{\ref{sec:algorithms}}).

Additional tests were conducted to measure the efficiency of the DIF filter under different parameters. The purpose of these experiments was to determine how the filtration efficiency is affected by the chosen parameters. The results indicate that updating inactive subareas has a negligible effect on\remove{ the} filtering efficiency. This becomes significant only when the noise intensity and motion in the observed scene are\remove{ both} very small, such as in the case of a static sensor. The AUROC values for the \textit{Street} dataset were 0.959 and 0.937 with and without the update of inactive areas, respectively, at an noise intensity of \SI{0.0038}{\Hz/px}. For the \textit{Corn} dataset, the AUROC values were 0.999 and 0.975, respectively. The differences between the AUROC ratios were around 0.001 for noise intensities \replace{above}{greater than} \SI{0.01}{\Hz/px}.
Table \ref{tab:AUROC_parameters} \replace{demonstrates}{shows} the average AUROC for different parameters of the DIF algorithm and noise levels (both artificial and recorded).

\begin{table}[!t]
	\centering
	\caption{Average AUROC values for all datasets for different parameters of the DIF algorithm.}
	\begin{tabular}{| c | c | c | c |}
        \multicolumn{1}{c}{} & \multicolumn{3}{c}{\textit{Scale}} \\
		\hline 
		  \textit{Update factor} & 8   & 16 & 32  \\	
		\hline
		0.125     & 0.8966  &  0.9070  &  0.8881 \\
		\hline
		0.25      & 0.9164  &  0.9071  &  0.8851 \\
		\hline
		0.5       & \textbf{0.9183}  &  0.9015  &  0.8731 \\
		\hline
	\end{tabular}
	\label{tab:AUROC_parameters}
\end{table}

Based on the data presented, the algorithm is most effective when dividing the sensor matrix into subareas with a size of \(8 \times 8\) and an update factor of 0.5. However, this size of subareas requires four times more data to be stored in memory than a size of \(16 \times 16\). For a size of \(16 \times 16\), the best result is achieved with an update factor of 0.25.\remove{ A} Performance deterioration is apparent for a size of \(32 \times 32\), as well as for a small size and a small update factor. \replace{Conversely}{In contrast}, larger subareas perform better with a smaller update factor.

\section{Hardware architecture}
\label{sec:hardwarearchitecture}
This section provides a detailed account of the hardware architecture that implements the DIF filter. It is divided into three subsections. The first describes the modifications that have been made to the architecture compared to the algorithm described in Section \mbox{\ref{sec:algorithms}}. The second subsection demonstrates how the algorithm has been realised using the reconfigurable resources of the FPGA chip. The third subsection presents an evaluation of the proposed solution.

\subsection{Modifications}
\label{ssec:modifications}
Several modifications were made to the algorithm in order to improve \replace{the architecture's efficiency}{the efficiency of the architecture} and reduce \replace{computational resources usage}{the utilisation of computational resources}. Efficiency is measured by the maximum frequency of operation, which also determines the throughput, or the number of processable events per unit of time. However, these modifications should not have a significant impact on the effectiveness of the filtering algorithm.

The initial modification involved the calculation of distances \(d_{11}\), \(d_{12}\), \(d_{21}\) and \(d_{22}\). To avoid excessive computational resource usage and\new{ a} negative impact on maximum operating frequency, the calculation\new{ of the L2 norm} was excluded. Instead, a read-only BRAM was used to memorise function values for \(\text{dx}\) and \(\text{dy}\), which can only take certain 16 values horizontally and vertically to neighbouring areas. \replace{Additionally}{Furthermore}, the distance values were rounded to two decimal bits, resulting in a precision reduced to 0.25.

A change was made to the calculation of coefficients \mbox{\(D_{11}\)}, \mbox{\(D_{12}\)}, \mbox{\(D_{21}\)} and \mbox{\(D_{22}\)} in \eqref{eq:DIF_SC}. The first step involves calculating the intermediate variables described in \eqref{eq:DIF_K}.

\begin{equation}
\begin{split}
    K_{11} &= I_{11} d_{11}\quad
    K_{12} = I_{12} d_{12}\\
    K_{21} &= I_{21} d_{21}\quad
    K_{22} = I_{22} d_{22}
\end{split}
\label{eq:DIF_K}
\end{equation}

\new{Then} a precision reduction of the calculation is\remove{ then} performed for these coefficients. The bits discarded include those corresponding to the fractional part resulting from multiplication by \(d\). We tested the possibility of discarding the last 7 to 10 bits.
However, reducing the coefficients could potentially decrease their value to 0. Equations \eqref{eq:DIF_SC} and \eqref{eq:DIF_multiply} indicate that this operation would also reduce the weight of neighbouring areas to 0. If this occurred in at least 2 of the 4 areas considered, both the \mbox{\(\Delta T_c\)} and the \mbox{\(F_c\)} variables would be 0, resulting in the rejection of the processed event. Therefore, a bottom saturation check was added to prevent such a case. If any of the coefficients \(K\) had a value of 0 after reduction, it was assigned a value of 1. Additionally, the maximum value was reduced by introducing saturation, which decreased the required number of bits. We checked the possibility of resizing to 11, 12, or 13 bits, which corresponded to saturation values of 2047, 4095, and 8191, respectively.

To decrease the number of multiplications required for the coefficients of \mbox{\(D\)} in \eqref{eq:DIF_SC}, they were arranged as follows: first, two multiplications were carried out \eqref{eq:DIF_KD}, followed by \eqref{eq:DIF_D}. This reduced the number of multiplications from 8 to 6.

\begin{equation}
    \begin{split}
        K_{d1} = K_{12} K_{21}\quad
        K_{d2} = K_{11} K_{22}
    \end{split}
\label{eq:DIF_KD}
\end{equation}

\begin{equation}
    \begin{split}
        D_{11} &= K_{d1} K_{22}\quad
        D_{12} =  K_{d2} K_{21}\\
        D_{21} &= K_{d2} K_{12}\quad
        D_{22} =  K_{d1} K_{11}
    \end{split}
\label{eq:DIF_D}
\end{equation}

The approach to handling boundary conditions has also been modified. If an event occurs in the red section (and therefore has only one adjacent area), as shown in Figure \ref{sfig:Borders}, subsequent calculations will assume that all the areas considered have the same timestamp and interval. Conversely, if the occurrence takes place in the green section, the read magnitudes are duplicated vertically or horizontally (depending on whether it is the horizontal or vertical edge of the matrix). This approach would be disadvantageous for general-purpose processor computing, but for FPGAs it allows the same computational resources to be used regardless of the position of the event.
A final simplification was the rounding of\new{ the} intervals and timestamps when updating\new{ the} areas to integer values.

The impact of the aforementioned modifications on filtering efficiency was assessed using the datasets \replace{outlined}{described} in Section \ref{sec:datasets}. Tables \ref{tab:AUROC_m3generatedNoise} and \ref{tab:AUROC_m3recordedNoise} present a comparison of the average AUCROC values between the original DIF method and the FPGA-adapted method. The former table displays the \replace{outcomes}{results} for artificial noise, while the latter presents the results for recorded noise.
The values were obtained for discarding the last 8 bits and reducing the size by saturation to 12 bits for\new{ the} variables \(K\) from \eqref{eq:DIF_K}.

\begin{table}[!t]
\centering
\caption{Average AUROC values for datasets with generated noise.}
\begin{tabular}{| c | c | c |}
    \hline 
    \textit{Dataset} & DIF   & DIF hardware  \\	
    \hline
    \textit{101}     & 0.946 & 0.946\\	
    \hline
    \textit{Library} & 0.863 & 0.863  \\
    \hline
    \textit{Model}   & 0.886 & 0.885\\
    \hline
    \textit{RIS}     & 0.856 & 0.856   \\
    \hline
    \textit{Corn}    & 0.999 & 0.999   \\
    \hline
    \textit{Street}  & 0.844 & 0.844 \\
    \hline
    \textit{Dancing} & 0.960 & 0.960  \\
    \hline
    \textit{People}  & 0.900 & 0.900 \\
    \hline
    Average          & 0.907 & 0.907 \\
    \hline
\end{tabular}
\label{tab:AUROC_m3generatedNoise}
\end{table}

\begin{table}[!t]
\centering
\caption{Average AUROC values for datasets with recorded noise}
\begin{tabular}{| c | c | c |}
    \hline 
    \textit{Dataset} & DIF   & DIF hardware  \\	
    \hline
    \textit{101}     & 0.942 & 0.942\\	
    \hline
    \textit{Library} & 0.873 & 0.874  \\
    \hline
    \textit{Model}   & 0.888 & 0.888\\
    \hline
    \textit{RIS}     & 0.855 & 0.855   \\
    \hline
    \textit{Corn}    & 0.999 & 0.999   \\
    \hline
    \textit{Street}  & 0.856 & 0.857 \\
    \hline
    \textit{Dancing} & 0.961 & 0.961  \\
    \hline
    \textit{People}  & 0.896 & 0.897 \\
    \hline
    Average          & 0.909 & 0.909 \\
    \hline
\end{tabular}
\label{tab:AUROC_m3recordedNoise}
\end{table}

It can be inferred from the given values that the suggested modifications had little to no impact on the filtration efficiency.

\subsection{Hardware realisation}
\label{ssec:hardwarerealisation}
\begin{figure}
    \centering
    \includegraphics[width=\linewidth]{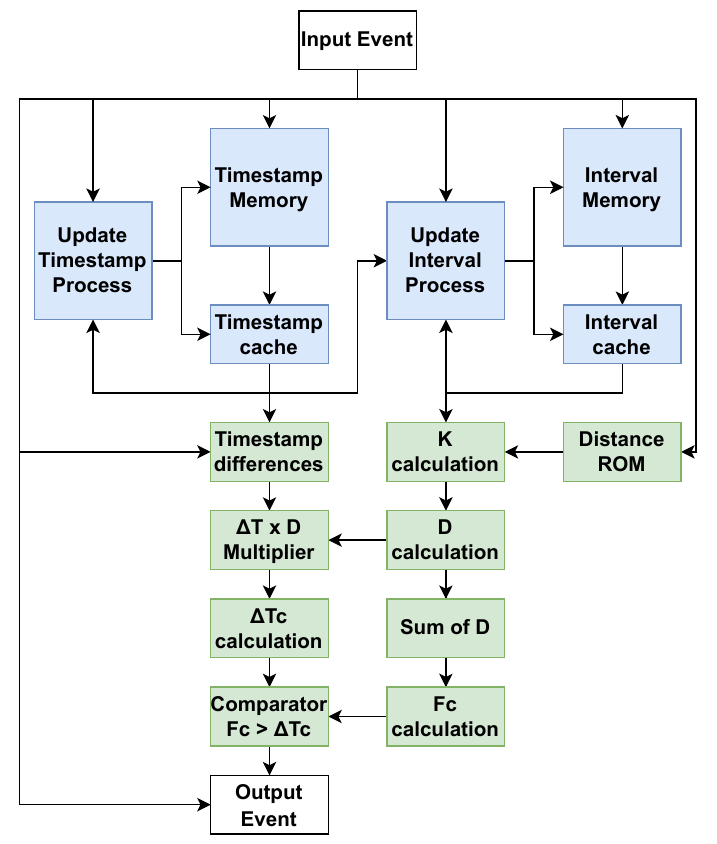}
    \caption{
    A simplified diagram of the architecture designed to realise the described DIF filtering algorithm. The blue elements update the area features, while the green elements interpolate the features and calculate the result.}
    \label{fig:Architecture}
\end{figure}

A simplified diagram of the proposed architecture is shown in Figure \ref{fig:Architecture}.
The blue elements are responsible for updating the features of the areas, while the green elements are responsible for interpolating the features and calculating the result. Some elements have been omitted from the figure for clarity, such as the updating of inactive subareas or the modules responsible for signal synchronisation.

The architecture was implemented using the \textit{SystemVerilog} hardware description language with the following parameters:
\begin{itemize}
    \item data width: 64 bits -- number of bits per input event,
    \item scale: 16 -- size of the square areas into which the sensor matrix has been divided (power of 2 in the current implementation),
    \item height: 720 -- height of sensor array,
    \item width: 1280 -- sensor array width,
    \item update offset: 2 -- by how many bits to shift the new area features (indirectly \(u\) in \eqref{eq:timestamp} and \eqref{eq:interval}),
    \item filter length: 200 -- \(F_L\) parameter in \eqref{eq:DIF_multiply}.
\end{itemize}
To assign an incoming event to the corresponding area, its coordinates are first subjected to a bit shift. The least significant bits of these coordinates, which are the remainder of\new{ the} dividing by the scale parameter, are then used to calculate the coordinates of all four areas in the vicinity of the event.
The parameters of these areas, namely, 4 timestamps and 4 intervals, are then read out. To maximise the throughput of the designed architecture, values were stored in parallel BlockRAMs, allowing\remove{ for} the parameters of adjacent areas to be read out in a single clock cycle. One port was used for reading values and the other for writing, with the same signals connected to the ports responsible for writing data to ensure data consistency between the memories responsible for timestamps and those responsible for intervals.

\begin{figure}
    \centering
    \includegraphics[width=0.8\linewidth]{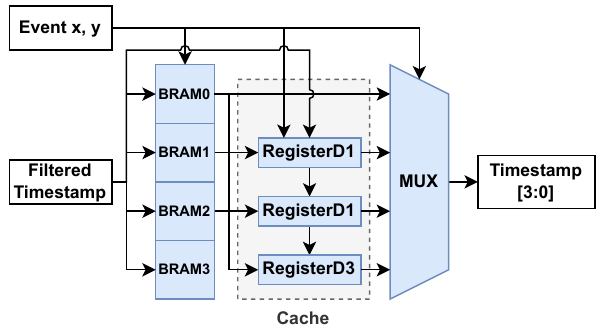}
    \caption{\new{Architecture of the BRAM memories.}}
    \label{fig:BRAM}
\end{figure}

The proposed architecture requires cache memory for the BRAMs described \replace{earlier}{above} to process events in each successive clock cycle. This is because writing updated data takes one clock cycle, while reading data takes two clock cycles (for the adopted configuration, chosen to maximise frequency). The data read-out refers to the memory state before three clock cycles. If any other event occurred during this time that affected the parameters of the read areas, those changes would not be taken into account. After reading the data, the system checks whether the coordinates of the read areas match the coordinates that were modified in the last three clock cycles. If they do, the read parameters are replaced with those that were sent to be written into memory.
\new{The architecture of the BRAM memories is shown in Figure \mbox{\ref{fig:BRAM}}.}

\replace{Furthermore}{In addition}, a matrix was created to store data on whether any events were reported in an area during the specified time period. This matrix is read and reset during a global update of inactive areas. It was realised using registers.

\begin{figure}
    \centering
    \includegraphics[width=0.8\linewidth]{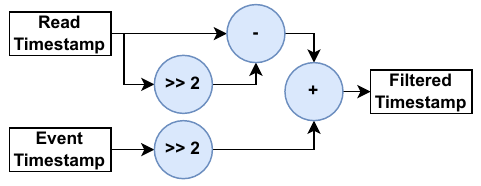}
    \caption{\new{Architecture for calculation of the filtered timestamp.}}
    \label{fig:FilterTimestamp}
\end{figure}

\begin{figure}
    \centering
    \includegraphics[width=0.8\linewidth]{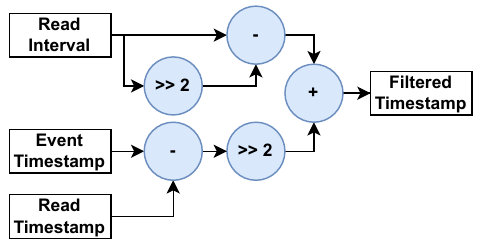}
    \caption{\new{Architecture for calculation of the filtered interval.}}
    \label{fig:FilterInterval}
\end{figure}

A new timestamp and interval are calculated based on the data read from memory and the timestamp of the currently processed event. These values are then written to the appropriate location in all BlockRAM memories.
\new{Architecture for calculation of the filtered timestamp, that will be written into the BRAM memory, is shown in Figure \mbox{\ref{fig:FilterTimestamp}}. The architecture for the calculation of the filtered interval is shown in Figure \mbox{\ref{fig:FilterInterval}}.}

After processing all events in a time interval, \replace{the updating of inactive areas}{updating inactive areas} \replace{commences}{begins}. The data from the activity registers \replace{of}{for} each area \replace{is}{are} read\remove{ out} one by one, and \replace{a}{the} timestamp and interval update\remove{s} \replace{is}{are} performed for the inactive areas.

Concurrent with the aforementioned process, calculations are conducted to determine whether an event should be rejected. They \replace{commence}{begin} by addressing the edge cases according to the modification outlined in Section \mbox{\ref{ssec:modifications}}.

The event's timestamp is subtracted from the timestamps of the neighbouring areas read from BlockRAM. The result is then truncated to 24 bits.

\begin{figure}
    \centering
    \includegraphics[width=0.8\linewidth]{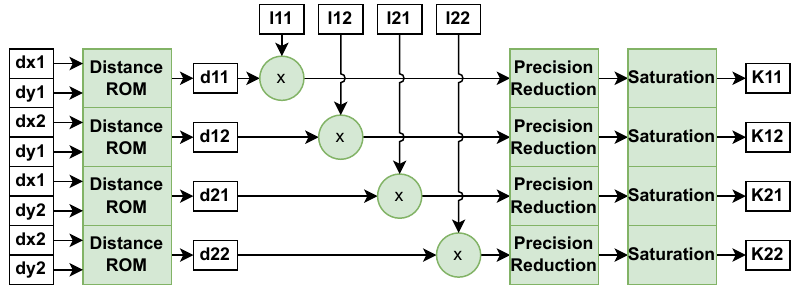}
    \caption{\new{Architecture for calculation of the distance to the subareas and calculations of \mbox{\(K\)} from \mbox{\eqref{eq:DIF_K}}.}}
    \label{fig:Distance}
\end{figure}

The values \(\text{dx}_1\), \(\text{dx}_2\), \(\text{dy}_1\) and \(\text{dy}_2\) are determined based on the last coordinate bits of the processed event, indicating the vertical and horizontal distances to neighbouring areas. The distance values are then read from the BlockRAM. It is important to note that, in this case, \replace{4}{four} memory banks with the same content were used to determine the distance for each area simultaneously.
\new{Scheme of the architecture realising these operations is shown in Figure \mbox{\ref{fig:Distance}}.}

The \replace{architecture's next element involved calculating}{next element of the architecture involved the calculation of the} factors \(K\) from \eqref{eq:DIF_K}. Multiplication was performed using DSP48 units, followed by precision reduction by truncating the last 8 bits.\new{ Then} saturation was\remove{ then} performed, and the number of bits was reduced to 12.
\new{Scheme of these operations is also shown in Figure \mbox{\ref{fig:Distance}}.}
The \replace{coefficients calculated}{calculated coefficients} were passed\new{ on} to the next set of multipliers. They calculated the \mbox{\(K_d\)} quantities using \eqref{eq:DIF_KD} and then next multipliers calculated the \mbox{\(D\)} values using \eqref{eq:DIF_D}.
The values of \mbox{\(D_{11}\)} and \mbox{\(D_{12}\)} were summed, as were the values of \mbox{\(D_{21}\)} and \mbox{\(D_{22}\)}. These two sums were then added together to obtain the final value of \(D_{sum}\). This value was then multiplied by the filter length parameter to determine \(F_c\), as \replace{per}{in} the first in \eqref{eq:DIF_multiply}.

Furthermore, the corresponding previously calculated timestamp differences were used to multiply \replace{with the}{by} \(D\). The \mbox{\(\Delta T_c\)} from \eqref{eq:DIF_multiply} value was then determined by a two-level summation.
The quantities \(F_c\) and \mbox{\(\Delta T_c\)} were compared. The delayed processed event and the final comparison result were transmitted to the module output.

\subsection{Architecture evaluation}
\label{ssec:ArchitectureEvaluation}
The hardware architecture described in the previous subsection was implemented using \textit{Vivado 2022.2} by \textit{AMD Xilinx}.
\remove{The }Synthesis was carried out for an \textit{Enclustra Mercury+ XU9} card, which includes a \textit{Zynq Ultrascale+ XCZU7EV-2FBVB900I} chip, and a \textit{Mercury+ ST1} base card.
\remove{The corresponding image can be found in Figure \mbox{\ref{fig:Mercury}}.}


To validate the FPGA architecture, a dataset was processed and compared with the software model. Initially, a behavioural simulation was used to check for correctness by processing the same events in both the simulation and\new{ the} software model. Consistent results were obtained.

To test the architecture while running on the FPGA, 8 BRAM banks were prepared, each containing events from a time interval of \SI{1}{\ms}, totalling 29031 events. The data from the banks were read one by one and transferred to the designed architecture. Inactive areas were globally updated between processing data from successive banks. \replace{Additionally}{In addition}, an IP Core \textit{ILA} (\textit{Integrated Logic Analyser}) was added inside the filter module. It allows the FPGA's signals' values to be checked during operation. The results were compared to a reference software model, achieving \replace{full}{complete} consistency.
However, it was not possible to check the operation with an event camera connected directly to the FPGA chip due to the output USB interface. This would have required support from the processor system with the operating system. However, compiling event camera drivers for the \textit{Petalinux} operating system, which is dedicated to\new{ the} \textit{Zynq} chips, is not possible, as \replace{no}{the} source code is\new{ not} publicly available. This issue might be solved in the future using the new \textit{Kria KV260 starter kit} released by \textit{Prophese}.

\new{A complete system that allows processing data directly from the event camera will certainly require more reconfigurable resources. A significant portion of this will have to be devoted to just receiving and communicating with the sensor. Higher resource consumption may in turn lead to poorer placement of the logical resources of the proposed architecture, which will reduce its efficiency. Depending on the format of the received data, there may also be a need to preprocess it so that a timestamp and horizontal and vertical coordinates are transmitted to the filtering module itself. The bottleneck of the system can also be outside of the proposed filter architecture, resulting in a reduction in maximum frequency and consequently a reduction in throughput.}

Table \ref{tab:zasoby} shows the resource utilisation for a resolution of \(1280 \times 720\) and \(640 \times 480\), and the maximum achievable operating frequencies of \replace{\mbox{\SI{319.69}{\MHz}}}{\mbox{\SI{312.70}{\MHz}}} and \replace{\mbox{\SI{409.33}{\MHz}}}{\mbox{\SI{400.32}{\MHz}}}, respectively.\new{ The resources specified in the table are: Look-Up Table (LUT), Look-Up Table Random Access Memory (LUTRAM), Flip-Flop (FF), Block Random Access Memory (BRAM), Global Clock Buffer (BUFG) and Digital Signal Processing (DSP) Block. LUTRAMs and BRAMs are used to store data. LUTs and FFs are used for implementing logic functions and simple computations. DSP blocks are used for more complex arithmetic operations, such as multiply-accumulate (MAC). BUFG ensures a consistent and low-skew distribution of the clock signal across the FPGA fabric.} It is important to note that the architecture cannot process events during the global update, which affects throughput. The global update length increases with resolution \replace{due to}{because of} the need to check more areas. To calculate the maximum processing speed, it is necessary to assume the interval between the update processes of inactive areas. The time required for the update depends on the number of areas to be checked and the \replace{chip frequency}{frequency of the chip}. For the designed architectures, it takes \replace{\mbox{\SI{11.28}{\us}}}{\mbox{\SI{11.53}{\us}}} and \replace{\mbox{\SI{3.78}{\us}}}{\mbox{\SI{3.01}{\us}}}, respectively. For updates every \SI{20}{\ms}, the throughput is reduced to \replace{\mbox{\SI{319.51}{MEPS}}}{\mbox{\SI{312.52}{MEPS}}} and \replace{\mbox{\SI{409.26}{MEPS}}}{\mbox{\SI{400.26}{MEPS}}}. In contrast, for updates every \SI{2}{\ms}, the values are \replace{\mbox{\SI{317.90}{MEPS}}}{\mbox{\SI{310.90}{MEPS}}} and \replace{\mbox{\SI{409.26}{MEPS}}}{\mbox{\SI{399.72}{MEPS}}}.
\new{The \textit{Vivado} estimation of the dynamic and static power for the \mbox{\(1280 \times 720\)} resolution were equal to \mbox{\SI{0.983}{W}} and \mbox{\SI{0.600}{W}}, respectively. For the \mbox{\(640 \times 480\)} resolution, these values were \mbox{\SI{0.708}{W}} and \mbox{\SI{0.597}{W}}, respectively.}
\new{Latency of the proposed architecture was equal to 30 clock cycles. This means that for the frequencies achieved, a result for the received event is calculated after \mbox{\SI{96}{\ns}} and \mbox{\SI{75}{\ns}}, respectively. It should be noted that if an event is received during the global update, it needs to be stored and processed after the update is finished. This can result in an increase in latency to the duration of the global update, which is \mbox{\SI{11.53}{\us}} and \mbox{\SI{3.01}{\us}}, respectively.}

\begin{table}[!t]
\centering
\caption{Resource utilisation of the proposed architecture.}
\begin{tabular}{| c | c | c |}
    \hline 
    {Resource} & {\(640 \times 480\)}   & \(1280 \times 720\)  \\	
    \hline
    LUT      & \replace{4709 (2.04\%)}{5323 (2.31\%)} & \replace{8778 (3.81\%)}{10023 (4.35\%)}\\
    \hline
    LUTRAM   & \replace{456 (0.45\%)}{516 (0.51\%)} & \replace{456 (0.45\%)}{518 (0.51\%)}  \\
    \hline
    FF       & \replace{6134 (1.33\%)}{6637 (1.44\%)} & \replace{8582 (1.86\%)}{9243 (2.01\%)}\\
    \hline
    BRAM     & 19.5 (6.25\%) & 57.5 (18.43\%)   \\
    \hline
    BUFG     & 2 (0.37\%) & 2 (0.37\%)   \\
    \hline
    DSP      & 22 (1.27\%)  & 22 (1.27\%) \\
    \hline
\end{tabular}
\label{tab:zasoby}
\end{table}

An analysis was conducted to determine the main reason for the limitation of the maximum operating frequency, particularly for higher resolution. The implementation results revealed that the longest critical paths were associated with the table containing area activities stored in\new{ the} registers. Therefore, moving this table from the registers to BlockRAM could potentially increase the operating frequency of the architecture.

However, according to the tests described in Section \ref{sec:evaluation}, updating inactive areas has a negligible effect on filtering results, particularly at low noise rates. We checked the resource utilisation and the achievable operating frequencies for an architecture without this element, as shown in Table \ref{tab:zasoby_nGU}. \replace{The maximum operating frequencies were \mbox{\SI{445.83}{\MHz}} and \mbox{\SI{469.04}{\MHz}}, respectively.}{The maximum throughputs were \mbox{\SI{403.39}{MEPS}} and \mbox{\SI{428.45}{MEPS}}, respectively.} \new{In this case the \textit{Vivado} estimation of the dynamic and static power for the \mbox{\(1280 \times 720\)} resolution were equal to \mbox{\SI{0.830}{W}} and \mbox{\SI{0.599}{W}}, respectively. For the \mbox{\(640 \times 480\)} resolution, these values were \mbox{\SI{0.616}{W}} and \mbox{\SI{0.597}{W}}, respectively.}
\new{Latency in this case was equal to \mbox{\SI{74}{\ns}} and \mbox{\SI{70}{\ns}}, respectively.}

\begin{table}[!t]
\centering
\caption{Resource utilisation of the proposed architecture without global update.}
\begin{tabular}{| c | c | c |}
    \hline 
    {Resource} & {\(640 \times 480\)}   & \(1280 \times 720\)  \\
    \hline
    LUT      & \replace{2859 (1.23\%)}{3079 (1.34\%)} & \replace{3059 (1.33\%)}{3357 (1.46\%)}\\	
    \hline
    LUTRAM   & \replace{448 (0.44\%)}{507 (0.50\%)} & \replace{448 (0.44\%)}{509 (0.50\%)}  \\
    \hline
    FF       & \replace{4824 (1.05\%)}{5306 (1.15\%)} & \replace{5105 (1.11\%)}{5457 (1.18\%)}\\
    \hline
    BRAM     & 19.5 (6.25\%) & 57.5 (18.43\%)   \\
    \hline
    BUFG     & 2 (0.37\%) & 2 (0.37\%)   \\
    \hline
    DSP      & 22 (1.27\%)  & 22 (1.27\%) \\
    \hline
\end{tabular}
\label{tab:zasoby_nGU}
\end{table}

In this case, the maximum frequency is equal to the throughput because a new event can be processed in each successive clock cycle. Additionally, there is no need to pause the data stream during a global update.

The \replace{obtained results}{results obtained} confirm that using an array of active areas in the form of registers for a global update process has a significant negative impact on the maximum clock frequency of the designed architecture. To achieve higher operating frequencies while retaining the update process, a possible solution would be to relocate the activity array from\new{ the} registers to BlockRAM. However, it is important to note that updating inactive areas globally only improves filtering results for very low noise intensities. Despite this, all tested architectures achieved very high throughput.

Based on the \replace{presented results}{results presented}, the algorithm implemented by the hardware architecture achieved an average AUROC value of 0.908 across the \replace{used datasets}{datasets used}, the same as the baseline method.

\section{Comparison with SOTA algorithms}
\label{sec:comparison}
Comparing the proposed solution with other state-of-the-art algorithms is a complex task. To ensure a fair evaluation, all solutions must be tested using the same datasets. 

Many \replace{methods based on neural networks}{neural network-based methods} often require architectural changes to process events from other sensors or resolutions. Direct comparison of these filtering algorithms would require\new{ the} reimplementation and training\new{ of} neural network-based solutions, which is a time-consuming process prone to implementation errors that could negatively impact their evaluation.
The filter and architecture proposed were designed to process high-resolution sensor events. Testing them on datasets with a small resolution, like \(346 \times 260\), could adversely affect their performance, as this would require very small subareas.

The comparison is divided into two parts. In the first, the filtering efficiency of the proposed methods with two sets of parameters is determined for 10-second sequences from the proposed dataset. The results obtained are then compared with those of the nearest-neighbour method (NNb) and its generalised version, the spatiotemporal correlation filter (STCF). The objective of this section is to present the results of the proposed methods for a large set, increasing their reliability. The second part presents a comparison of the performance of the proposed algorithm with two state-of-the-art solutions: EDnCNN and AEDNet. However, the dataset in the second part was very limited due to the long inference time of the neural networks. Additionally, the DVSCLEAN dataset was used in this comparison in order to provide\new{ a} comparison with a dataset that is already publicly available. This part aims to demonstrate how the results achieved position themselves against highly advanced event data filtering methods.

In the first part, similar \replace{as in}{to} the paper \mbox{\cite{guo2022low}}, the STCF algorithm was used as a reference \replace{for}{to} evaluate the proposed methods. This method is a development of the fourth method described in \cite{czech2016evaluating}, which in turn was a development of the \replace{nearest neighbour}{NNb} method. The \replace{nearest neighbour}{NNb} method assumes that there must be at least one other event in the vicinity of the event being processed. In the work \cite{czech2016evaluating}, this assumption was modified so that there should be at least two events. In the STCF method, the number of pixels \replace{from}{in} the vicinity for which an event is to be registered is a parameter \(N\). For \(N\) equal to 1, the method is equivalent to the \replace{nearest neighbour}{NNb} method. For \(N\) equal to 2, it corresponds to the method proposed in \cite{czech2016evaluating}. Tests were \replace{conducted}{carried out} on the provided datasets to determine the best\new{ value of} \(N\)\remove{ value}. The results are presented in Table \ref{tab:AUROC_STCFparameters}, which indicates that the algorithm performed best with a parameter \(N\) equal to 2. However, the improvement over the basic version of the \replace{nearest neighbour}{NNb} algorithm was only marginal.

\begin{table}[!t]
	\centering
	\caption{Average AUROC values for all datasets for different parameters of the STCF algorithm.}
	\begin{tabular}{| c | c | c | c | c | c |}
        \hline
		  \textit{\(N\)} & 1   & 2 & 3 & 4 & 5  \\	
		\hline
		AUROC     & 0.913  &  \textbf{0.914}  &  0.897 & 0.870 & 0.836 \\
		\hline
	\end{tabular}
	\label{tab:AUROC_STCFparameters}
\end{table}

\begin{figure}
    \centering
    \includegraphics[width=\linewidth]{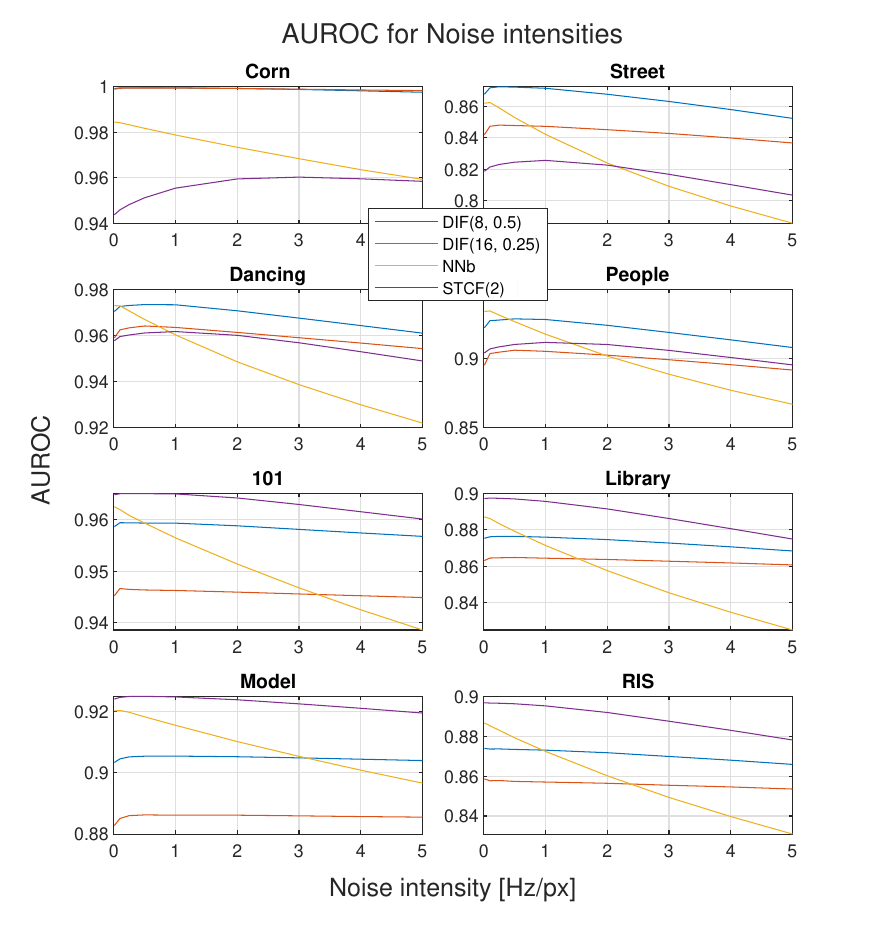}
    \caption{
    Filtering efficiency plotted against noise intensity for the test sequences.}
    \label{fig:AUROCnoiseIntensitiesDatasets_comparison}
\end{figure}

The DIF algorithm with scale parameters of 8 and 16, and update factors of 0.5 and 0.25, respectively, was used for comparison. The noise generated was used to plot\new{ the} AUROC graphs, which are shown in Figure \ref{fig:AUROCnoiseIntensitiesDatasets_comparison}. Each graph represents a different dataset. The results indicate that for datasets recorded with the event camera, the best performing filter is DIF(8, 0.5), followed by DIF(16, 0.25), with STCF(2) only surpassing the latter for the \textit{People} dataset. In contrast, STCF(2) is the most effective algorithm for \textit{V2E}-generated datasets, followed by DIF(8, 0.5).

\begin{figure}
    \centering
    \includegraphics[width=\linewidth]{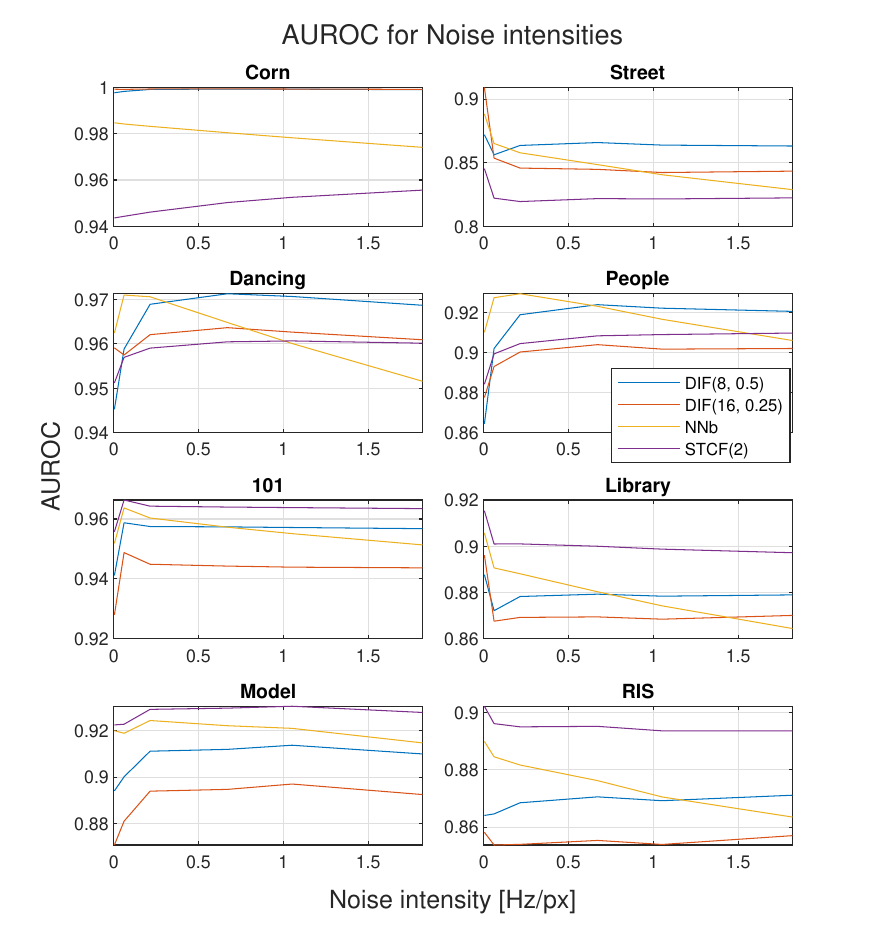}
    \caption{
    Plots showing the effectiveness of filtering against recorded noise intensity for the test sequences.}
    \label{fig:AUROCnoiseIntensitiesCombinedDatasets_comparison}
\end{figure}

Figure \ref{fig:AUROCnoiseIntensitiesCombinedDatasets_comparison} illustrates the comparison of noise recorded with an event sensor. The DIF algorithms outperformed the STCF method for\remove{ the} recorded sequences, while the STCF method performed better for the \textit{V2E}-generated sequences. When comparing Figures \ref{fig:AUROCnoiseIntensitiesDatasets_comparison} and \ref{fig:AUROCnoiseIntensitiesCombinedDatasets_comparison}, it is important to note that the range of noise intensity is smaller for the recorded ones. When considering the common interval of noise intensity, ranging from \SI{0.01}{\Hz/px} to \SI{1.81}{\Hz/px}, it is evident that the graphs in this range are highly comparable. This suggests that the generated noise accurately reflects the actual noise for the event cameras.

\begin{figure}
    \centering
    \includegraphics[width=0.8\linewidth]{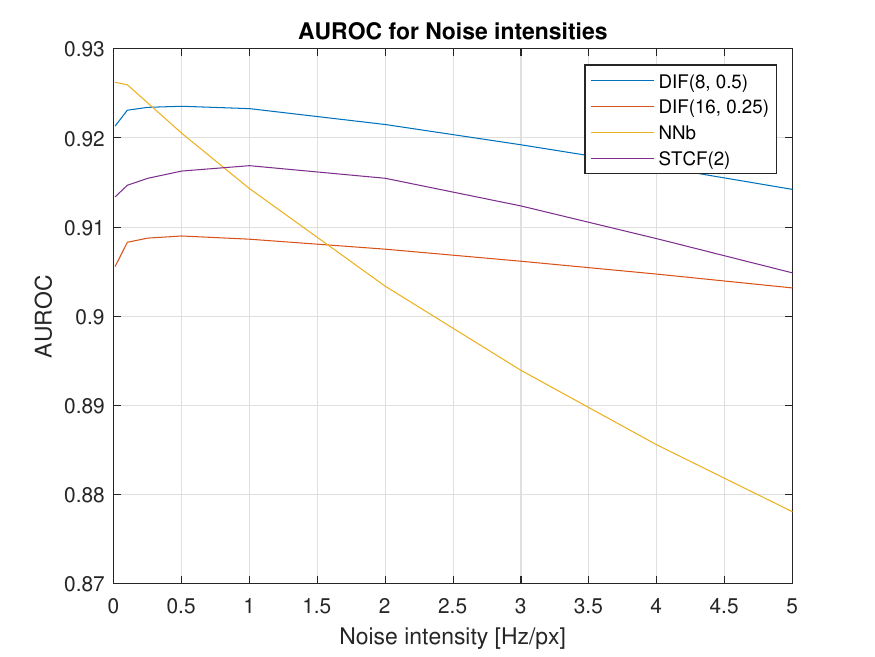}
    \caption{Comparison of \replace{filtering efficiency for various}{the average AUROC value over all test sequences for different} levels of generated noise.}
    \label{fig:AUROCnoiseIntensities_comparison}
\end{figure}

Figure \ref{fig:AUROCnoiseIntensities_comparison} combines all the graphs from Figure \ref{fig:AUROCnoiseIntensitiesDatasets_comparison} into one, showing the overall performance of the methods for different intensities of generated noise. The best performing method in this comparison was DIF(8, 0.5), followed by\remove{ the} STCF(2). It \replace{is also noteworthy}{should also be noted} that the NNb algorithm performs best among the compared methods for low noise intensity (up to \SI{0.25}{\Hz/px}), but loses effectiveness much faster as the intensity increases compared to the other algorithms.
\new{The differences between the highest and lowest values of the AUROC were also calculated to assess the stability of the filters at different levels of generated noise. For DIF(8, 0.5), DIF(16, 0.25), NNb and STCF(2) the decreases were equal to 1.01\%, 0.64\%, 5.20\% and 1.31\%, respectively. This shows that the proposed DIF method has a very stable performance over a wide range of noise intensities.}

\begin{figure}
    \centering
    \includegraphics[width=0.8\linewidth]{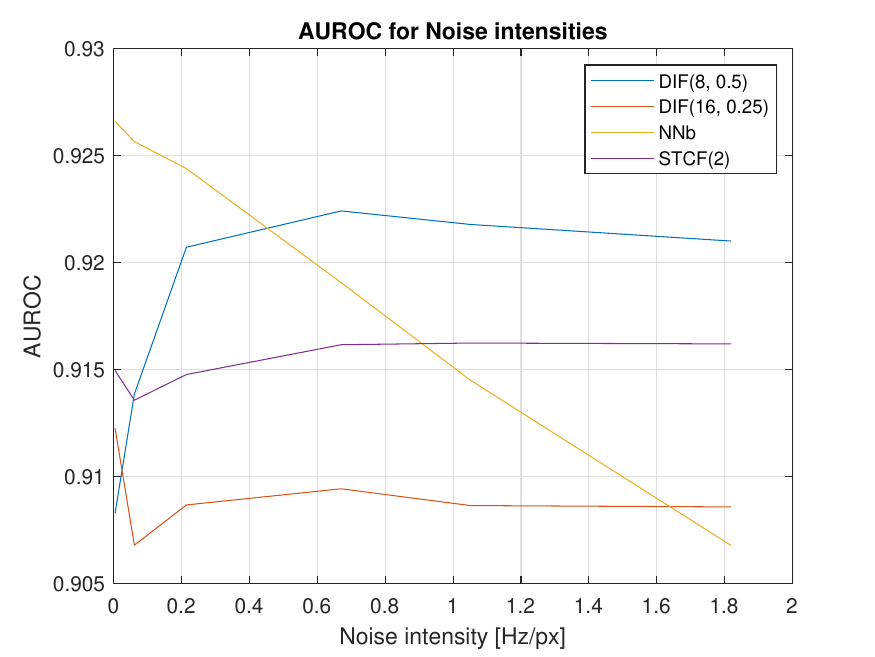}
    \caption{Comparison of \replace{filtering efficiency for various}{the average AUROC value over all test sequences for different} levels of recorded noise.}
    \label{fig:AUROCnoiseIntensitiesCombined_comparison}
\end{figure}

Figure \ref{fig:AUROCnoiseIntensitiesCombined_comparison} displays the same graph for recorded noise. The performance of the DIF(8, 0.5) algorithm was worse when the noise intensity was very low, but improved rapidly as the noise intensity increased. The earlier observation that the NNb algorithm performs best for low noise intensities but quickly loses effectiveness as the intensity increases is also confirmed.

Table \ref{tab:AUROC_averageComparison} shows a comparison of efficiency for all datasets and noise levels. The DIF(8, 0.5) algorithm performed the best, while the NNb and STCF(2) methods had very similar average efficiencies of 0.913 and 0.914, respectively. The DIF(16, 0.25) algorithm performed slightly worse, with an average efficiency of 0.908. \replace{It is worth noting}{The table also shows} that the hardware implementation of the DIF(16, 0.25) method achieved the same AUROC values on average, as previously mentioned in Section \ref{sec:hardwarearchitecture}.

It \replace{is worth noting}{should be noted} that even the proposed hardware implementation of the DIF algorithm with a scale parameter of 8 would use 8 times less memory than the NNb or STCF algorithms, which means that even at a resolution of \(1280 \times 720\) it would fit in the FPGA chip used. However, it would certainly have a lower throughput due to the greater dispersion of BlockRAM resources, which would have a negative impact on critical paths.

\begin{table}[!t]
\centering
\caption{Average AUROC values for\new{ all} datasets.}
\begin{tabular}{| c | c | c | c | c |}
    \hline 
    \textit{Dataset} & DIF & DIF & NNb & STCF  \\	
     & (8, 0.5) & (16, 0.25) &  & (2)  \\	
    \hline
    \textit{101}     & 0.957  &  0.944  &  0.955  &  \textbf{0.964}  \\	
    \hline
    \textit{Library} & 0.876  &  0.867  &  0.872  &  \textbf{0.895}  \\
    \hline
    \textit{Model}   & 0.906  &  0.887  &  0.915  &  \textbf{0.925}  \\
    \hline
    \textit{RIS}     & 0.870  &  0.856  &  0.870  &  \textbf{0.893}  \\
    \hline
    \textit{Corn}    & \textbf{0.999}  &  0.999  &  0.978  &  0.952  \\
    \hline
    \textit{Street}  & \textbf{0.865}  &  0.849  &  0.842  &  0.821  \\
    \hline
    \textit{Dancing} & \textbf{0.967}  &  0.961  &  0.958  &  0.958  \\
    \hline
    \textit{People}  & \textbf{0.917}  &  0.899  &  0.913  &  0.905  \\
    \hline
    Average          & \textbf{0.920}  &  0.908  &  0.913  &  0.914  \\
    \hline
\end{tabular}
\label{tab:AUROC_averageComparison}
\end{table}

\begin{figure}
    \centering
    \includegraphics[width=0.8\linewidth]{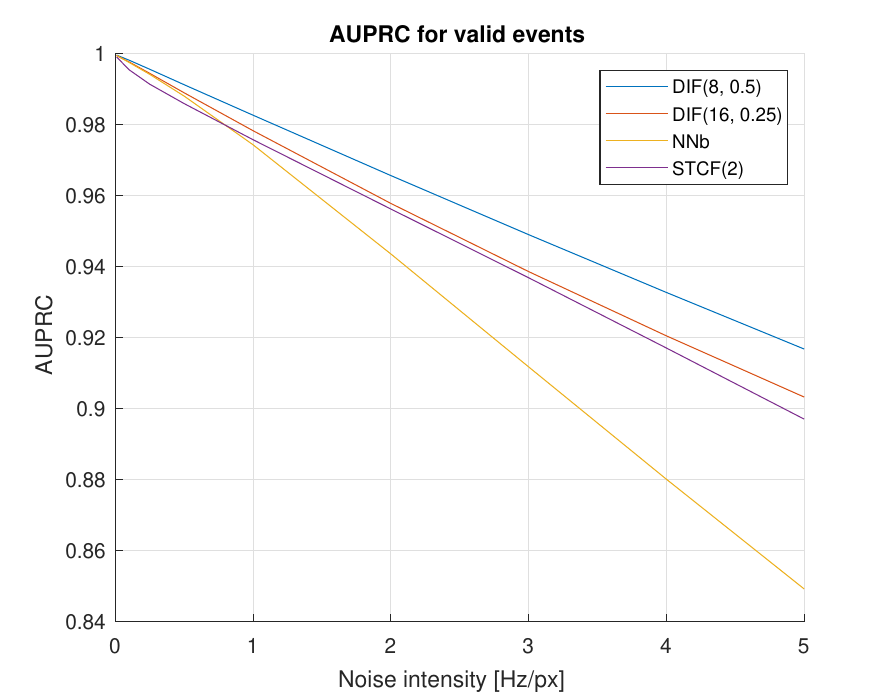}
    \caption{\new{Comparison of the average AUPRC value over all test sequences for different levels of generated noise.}}
    \label{fig:AUPRCnoiseIntensities_comparison}
\end{figure}

\new{To provide a better evaluation of the tested methods, an Area Under the Precision-Recall Curve (AUPRC) was also calculated. Precision measures what part of the samples predicted to be valid events are actually valid. Recall measures what proportion of all valid events were correctly predicted.
As with AUROC, a higher AUPRC value indicates a better filtering result for the class under consideration.
Whereas AUROC measures the ability of the model to discriminate between classes regardless of threshold and considers all examples, both positive and negative, AUPRC focuses on the quality of positive class detection and considers only the hits and misses associated with positive examples.}
\new{Figure \mbox{\ref{fig:AUPRCnoiseIntensities_comparison}} shows the average AUPRC results of all the test sequences for different levels of noise intensity of the generated noise. 
The best performing method in this comparison was DIF(8, 0.5), followed by DIF(16, 0.25). Also in this comparison, the NNb method achieves good results for small noise intensity (up to \mbox{\SI{0.5}{\Hz/px}}), but loses effectiveness much faster than other methods.}

\begin{figure}
    \centering
    \includegraphics[width=0.8\linewidth]{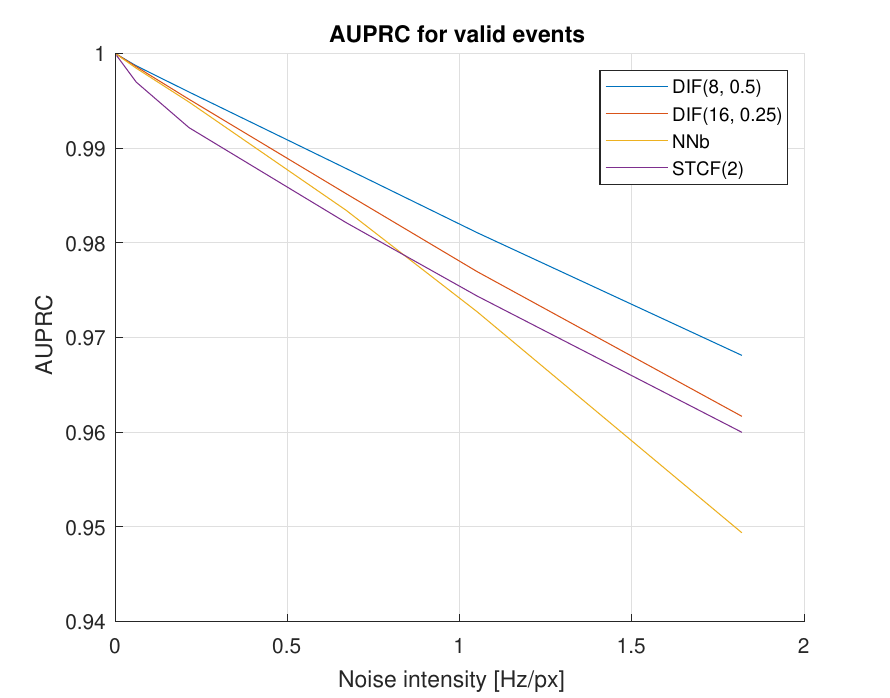}
    \caption{\new{Comparison of the average AUPRC value over all test sequences for different levels of recorded noise.}}
    \label{fig:AUPRCnoiseIntensitiesCombined_comparison}
\end{figure}

\new{The AUPRC was also compared for the recorded noise in Figure \mbox{\ref{fig:AUPRCnoiseIntensitiesCombined_comparison}}. The results obtained are very similar to those presented above for the generated noise.}

The second part of the tests was conducted to evaluate the efficacy of the proposed solutions compared to two SOTA algorithms: EDnCNN\mbox{\cite{baldwin2020event}} and AEDNet\mbox{\cite{fang2022aednet}}. These were chosen based on the high efficiency reported in the original papers and their compatibility with datasets of \mbox{\(1280 \times 720\)} resolution.
\replace{Additionally}{In addition}, the authors of both papers provided trained models of the networks.
\new{It was decided to use these provided model parameters as this allows a better comparison of the results obtained with those presented in the original work. It also avoids the risk of incorrectly training the artificial intelligence model used.}
A softmax layer is present at the end of both networks, which outputs a two-element vector indicating the probability that an event is a noise. The event classification threshold can be varied from 0 to 1 to produce an ROC curve, and from this the AUROC value can be calculated. Thus, the algorithms were compared regardless of the threshold chosen.

The initial comparison was conducted using the simulated subset of the \textit{DVSCLEAN} dataset\mbox{\cite{fang2022aednet}}.
The results\new{ for the validation set} are presented in Table \mbox{\ref{tab:AUROC_DVSCLEAN}}\new{ for the AUROC values and in Table \mbox{\ref{tab:AUPRC_DVSCLEAN}} for the AUPRC values}. The comparison considers the original DIF algorithm, the hardware architecture model, \replace{the nearest neighbor}{NNb}, AEDNet, and EDnCNN.
It should be acknowledged that this comparison \replace{is}{may be} biased in favour of the AEDNet algorithm, \replace{as the model provided was trained on at least a portion of this dataset}{since the provided model was trained on the  training subset of the DVSCLEAN dataset, and the validation set was used to select the most efficient parameters of the model}.
\replace{It is also noteworthy}{It should also be noted} that all the algorithms for this dataset yielded very good \replace{outcomes}{results}. This may indicate that the dataset is relatively straightforward and \replace{comprises}{includes} a limited number of complex scenarios.


\begin{table}[!t]
\centering
\caption{\new{Average AUROC values for the \textit{DVSCLEAN} dataset.}}
\begin{tabular}{| c | c | c | c | c | c |}
    \hline 
    \textit{Noise}  & DIF & Model & NNb & AED & EDn    \\
      & (16, 0.25) &  &  &  &     \\	
    \hline
    \textit{50\%}   & 0.973  &  0.969  &  0.978  & \textbf{0.994} & 0.980  \\	
    \hline
    \textit{100\%}  & 0.971  &  0.965  &  0.966  & \textbf{0.990} & 0.972 \\
    \hline
\end{tabular}
\label{tab:AUROC_DVSCLEAN}
\end{table}

\begin{table}[!t]
\centering
\caption{\new{Average AUPRC values for the \textit{DVSCLEAN} dataset.}}
\begin{tabular}{| c | c | c | c | c | c |}
    \hline 
    \textit{Noise}  & DIF(16, 0.25) & Model & NNb & AED & EDn    \\
      & (16, 0.25) &  &  &  &     \\	
    \hline
    \textit{50\%}   & 0.983  &  0.981  &  0.988  & \textbf{0.997} & 0.992  \\	
    \hline
    \textit{100\%}  & 0.968  &  0.964  &  0.962  & \textbf{0.991} & 0.981 \\
    \hline
\end{tabular}
\label{tab:AUPRC_DVSCLEAN}
\end{table}

\begin{figure}
    \centering
    \includegraphics[width=0.8\linewidth]{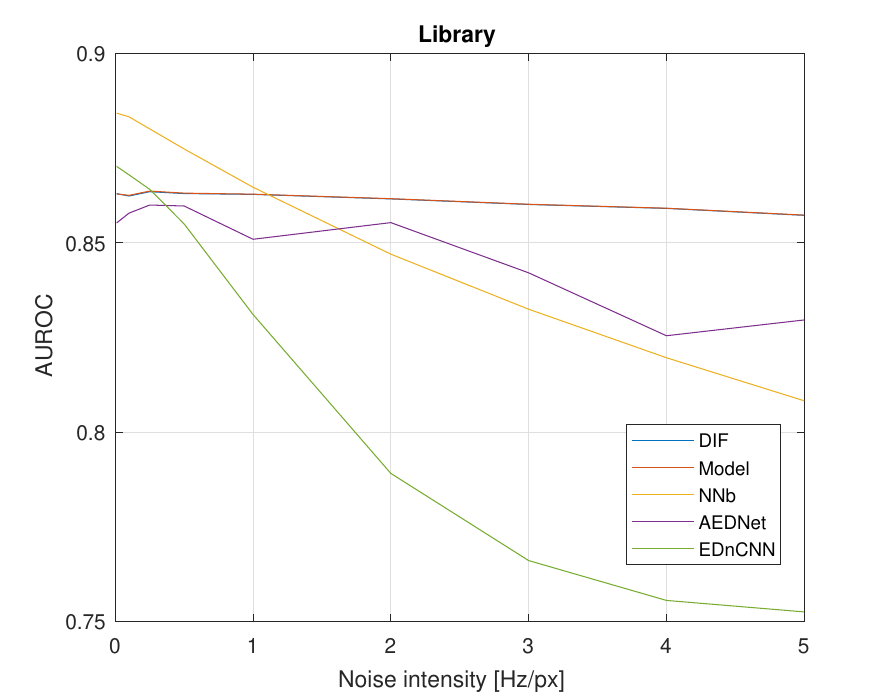}
    \caption{\new{AUROC} comparison of filtering algorithms for various levels of noise intensity\new{ for the \textit{Library} sequence.}}
    \label{fig:AUROC_Library_comparison}
\end{figure}

\begin{figure}
    \centering
    \includegraphics[width=0.8\linewidth]{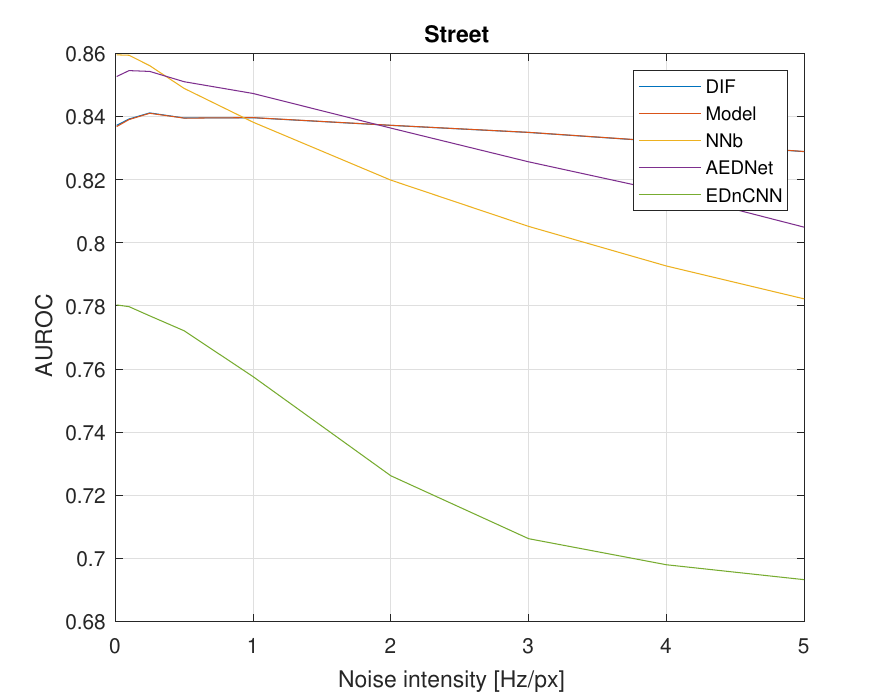}
    \caption{\new{AUROC} comparison of filtering algorithms for various levels of noise intensity\new{ for the \textit{Street} sequence.}}
    \label{fig:AUROC_Street_comparison}
\end{figure}

Subsequently, a test for a subset of our dataset was also carried out. To this end, the sequences \textit{Library} and \textit{Street} were selected, and\new{ the} data comprising \replace{one}{a} second were extracted from each.
The extensive limitation of the dataset was necessitated by the considerable computational complexity of the inference of algorithms based on neural networks. The sequences were used with varying noise intensity, spanning a range from \mbox{\SI{0.01}{\Hz/px}} do \mbox{\SI{5}{\Hz/px}}.
The AUROC values for the \textit{Library} sequence are presented in Figure \mbox{\ref{fig:AUROC_Library_comparison}} and Table \mbox{\ref{tab:AUROC_Library_comparison}}. \replace{Additionally}{In addition}, the table provides the mean AUROC value for each method and the percentage change in AUROC with increasing noise level. The percentage change was calculated as the difference between the maximum and minimum values divided by the maximum value.
The results for the \textit{Street} sequence are presented in Figure \mbox{\ref{fig:AUROC_Street_comparison}} and Table \mbox{\ref{tab:AUROC_Street_comparison}}.

\begin{table}[!t]
\centering
\caption{\new{AUROC values, mean and decrease for the \textit{Library} sequence.}}
\begin{tabular}{| c | c | c | c | c | c |}
    \hline 
    Noise & DIF & Model & NNb & AED & EDn \\
    (\SI{}{\Hz/px}) & (16, 0.25) & & & & \\
    \hline
    0.01  &  0.863  &  0.863  &  \textbf{0.884}  &  0.855  &  0.870  \\    
    \hline
    0.10  &  0.862  &  0.863  &  \textbf{0.883}  &  0.858  &  0.868  \\    
    \hline
    0.25  &  0.864  &  0.864  &  \textbf{0.880}  &  0.860  &  0.864  \\    
    \hline
    0.50  &  0.863  &  0.863  &  \textbf{0.875}  &  0.860  &  0.855  \\    
    \hline
    1.00  &  0.863  &  0.863  &  \textbf{0.865}  &  0.851  &  0.831  \\    
    \hline
    2.00  &  0.862  &  \textbf{0.862}  &  0.847  &  0.855  &  0.789  \\    
    \hline
    3.00  &  0.860  &  \textbf{0.860}  &  0.832  &  0.842  &  0.766  \\    
    \hline
    4.00  &  0.859  &  \textbf{0.859}  &  0.820  &  0.825  &  0.756  \\    
    \hline
    5.00  &  0.857  &  \textbf{0.857}  &  0.808  &  0.830  &  0.753  \\    
    \hline
    Average  &  0.861  &  \textbf{0.861}  &  0.855  &  0.848  &  0.817  \\    
    \hline
    Dec. [\%] &  \textbf{0.72}  &  0.74  &  8.59  &  4.01  &  13.52  \\    
    \hline
\end{tabular}
\label{tab:AUROC_Library_comparison}
\end{table}

\begin{table}[!t]
\centering
\caption{\new{Average AUROC values, mean and decrease for the \textit{Street} sequence.}}
\begin{tabular}{| c | c | c | c | c | c |}
    \hline 
    Noise & DIF & Model & NNb & AED & EDn \\
    (\SI{}{\Hz/px}) & (16, 0.25) & & & & \\
    \hline
    0.01  &  0.837  &  0.837  &  \textbf{0.860}  &  0.853  &  0.780  \\    
    \hline
    0.10  &  0.839  &  0.839  &  \textbf{0.859}  &  0.855  &  0.780  \\    
    \hline
    0.25  &  0.841  &  0.841  &  \textbf{0.856}  &  0.854  &  0.777  \\    
    \hline
    0.50  &  0.840  &  0.840  &  0.849  &  \textbf{0.851}  &  0.772  \\    
    \hline
    1.00  &  0.840  &  0.840  &  0.838  &  \textbf{0.847}  &  0.758  \\    
    \hline
    2.00  &  0.837  &  \textbf{0.837}  &  0.820  &  0.836  &  0.726  \\    
    \hline
    3.00  &  0.835  &  \textbf{0.835}  &  0.805  &  0.826  &  0.706  \\    
    \hline
    4.00  &  0.832  &  \textbf{0.832}  &  0.793  &  0.816  &  0.698  \\    
    \hline
    5.00  &  0.829  &  \textbf{0.829}  &  0.782  &  0.805  &  0.693  \\    
    \hline
    Average    &  0.837  &  0.837  &  0.829  &  \textbf{0.838}  &  0.743  \\    
    \hline
    Dec. [\%]  &  1.46  &  \textbf{1.44}  &  9.00  &  5.81  &  11.16  \\    
    \hline
\end{tabular}
\label{tab:AUROC_Street_comparison}
\end{table}

\remove{In order to ensure an accurate assessment of the \textit{EDnCNN} and \textit{AEDNet} methods, the efficiency calculation was only performed after 20\% of the data in a given set had been processed. This approach was employed to prevent the determined efficiency from being underestimated by the initialization of methods based on neural networks, when they lack neighbours in the analysed spatial-temporal environment.}

\begin{figure}
    \centering
    \includegraphics[width=0.8\linewidth]{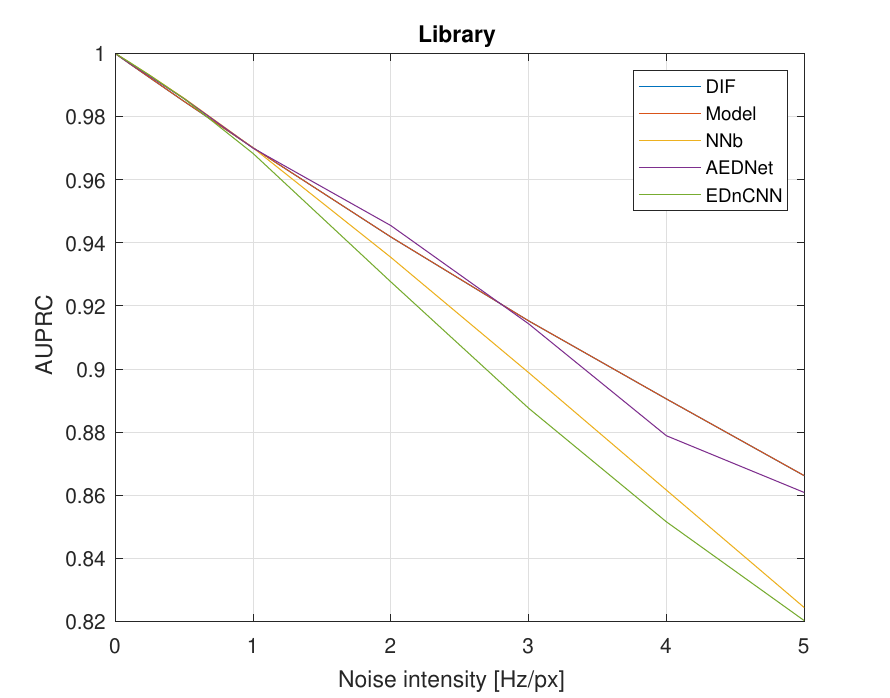}
    \caption{\new{AUPRC comparison of filtering algorithms for various levels of noise intensity for the \textit{Library} sequence.}}
    \label{fig:AUPRC_Library_comparison}
\end{figure}

\begin{figure}
    \centering
    \includegraphics[width=0.8\linewidth]{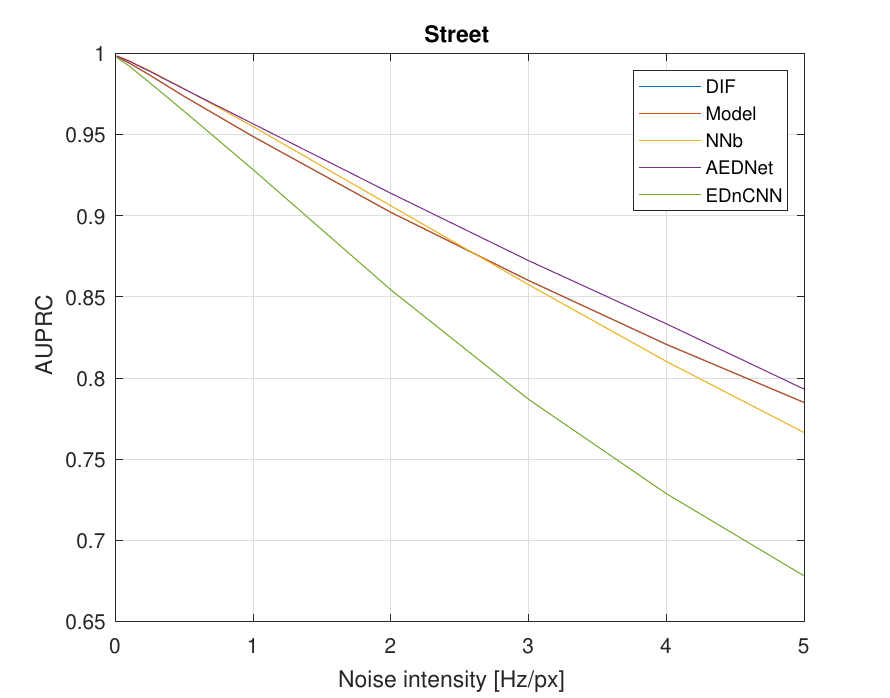}
    \caption{\new{AUPRC comparison of filtering algorithms for various levels of noise intensity for the \textit{Street} sequence.}}
    \label{fig:AUPRC_Street_comparison}
\end{figure}

\new{A similar comparison for the AUPRC values and the \textit{Library} sequence is presented in Figure \mbox{\ref{fig:AUPRC_Library_comparison}} and Table \mbox{\ref{tab:AUPRC_Library_comparison}}. AUPRC values for the \textit{Street} sequence are presented in Figure \mbox{\ref{fig:AUPRC_Street_comparison}} and Table \mbox{\ref{tab:AUPRC_Street_comparison}}.}

\begin{table}[!t]
\centering
\caption{\new{AUPRC values, mean and decrease for the \textit{Library} sequence.}}
\begin{tabular}{| c | c | c | c | c | c |}
    \hline 
    Noise & DIF & Model & NNb & AED & EDn \\
    (\SI{}{\Hz/px}) & (16, 0.25) & & & & \\
    \hline
    0.01  &  1.000  &  1.000  &  1.000  &  1.000  &  \textbf{1.000}  \\    
    \hline
    0.10  &  0.997  &  0.997  &  \textbf{0.997}  &  0.997  &  0.997  \\    
    \hline
    0.25  &  0.992  &  0.992  &  \textbf{0.993}  &  0.993  &  0.993  \\    
    \hline
    0.50  &  0.985  &  0.985  &  \textbf{0.986}  &  0.986  &  0.986  \\    
    \hline
    1.00  &  \textbf{0.970}  &  0.970  &  0.970  &  0.970  &  0.968  \\    
    \hline
    2.00  &  0.942  &  0.942  &  0.935  &  \textbf{0.945}  &  0.928  \\    
    \hline
    3.00  &  0.915  &  \textbf{0.915}  &  0.899  &  0.914  &  0.888  \\    
    \hline
    4.00  &  \textbf{0.891}  &  0.891  &  0.862  &  0.879  &  0.852  \\    
    \hline
    5.00  &  0.866  &  \textbf{0.866}  &  0.824  &  0.861  &  0.820  \\    
    \hline
    Mean  &  0.951  &  \textbf{0.951}  &  0.941  &  0.949  &  0.937  \\    
    \hline
    Dec. [\%] &  13.35  &  \textbf{13.35}  &  17.53  &  13.88  &  17.95  \\    
    \hline
\end{tabular}
\label{tab:AUPRC_Library_comparison}
\end{table}

\begin{table}[!t]
\centering
\caption{\new{AUPRC values, mean and decrease for the \textit{Street} sequence.}}
\begin{tabular}{| c | c | c | c | c | c |}
    \hline 
    Noise & DIF & Model & NNb & AED & EDn \\
    (\SI{}{\Hz/px}) & (16, 0.25) & & & & \\
    \hline
    0.01  &  0.998  &  0.998  &  \textbf{0.999}  &  0.998  &  0.998  \\    
    \hline
    0.10  &  0.994  &  0.994  &  \textbf{0.996}  &  0.995  &  0.992  \\    
    \hline
    0.25  &  0.987  &  0.987  &  \textbf{0.989}  &  0.989  &  0.982  \\    
    \hline
    0.50  &  0.974  &  0.974  &  \textbf{0.978}  &  0.978  &  0.964  \\    
    \hline
    1.00  &  0.949  &  0.949  &  0.955  &  \textbf{0.957}  &  0.928  \\    
    \hline
    2.00  &  0.902  &  0.902  &  0.906  &  \textbf{0.914}  &  0.854  \\    
    \hline
    3.00  &  0.860  &  0.860  &  0.858  &  \textbf{0.872}  &  0.787  \\    
    \hline
    4.00  &  0.821  &  0.821  &  0.810  &  \textbf{0.833}  &  0.729  \\    
    \hline
    5.00  &  0.785  &  0.785  &  0.766  &  \textbf{0.793}  &  0.678  \\    
    \hline
    Mean  &  0.919  &  0.919  &  0.917  &  \textbf{0.926}  &  0.879  \\    
    \hline
    Dec. [\%] &  21.36  &  21.36  &  23.25  &  \textbf{20.56}  &  32.03  \\    
    \hline
\end{tabular}
\label{tab:AUPRC_Street_comparison}
\end{table}

Several conclusions can be drawn from the comparisons made.
First, the average AUROC values for the DIF, NNb, and AEDNet methods are very close to each other, with a difference of less than 0.015. The only method that exhibits a notable discrepancy in this regard is EDnCNN, which exhibits a lower AUROC value of approximately 0.09 for the \textit{Street} sequence. However, it \replace{is noteworthy}{should be noted} that for the \textit{Library} sequence at low noise intensity, EDnCNN demonstrated superior performance compared to both the proposed methods and AEDNet.
\new{A similar observation can be made for the AUPRC comparison, where the average results for the DIF, NNb, and AEDNet methods are also very close to each other, while the EDnCNN method performs worse, especially for the Street sequence.}
Second, the designed filtering methods demonstrate efficacy in filtering events at varying levels of noise intensity. The reduction in the AUROC coefficient across a range of noise intensities from \mbox{\SI{0.01}{\Hz/px}} to \mbox{\SI{5}{\Hz/px}} is minimal, \replace{amounting to a mere}{which equates to just} 1.5\% for the sequences tested. A decrease of 6\% was observed for the AEDNet method. The largest decrease in efficiency was observed for the EDnCNN method, \replace{amounting to 16\%}{which is 13.5\%}.
\new{For AUPRC, the decrease in value with increasing interference intensity was much greater. For the \textit{Library} sequence, it was smallest for the DIF method at 13.35\%, but only slightly larger for the AEDNet method at 13.88\%. On the other hand, for the \textit{Street} sequence, the smallest decrease of 20.56\% was observed for the AEDNet method. In contrast, it was 21.36\% for the proposed method. Also for AUPRC, the largest decrease in efficiency with increasing interference intensity was observed for the EDnCNN method.}
Third, the computational times of the neural network-based methods are significantly longer than those of the proposed method. For the DIF algorithm, processing 18 test sequences containing a total of 63426038 events on a single Intel Xeon Platinum processor core took a total of 2 minutes and 11 seconds. 
For the algorithms \textit{EDnCNN} and \textit{AEDNet} running on 16 cores of the \textit{AMD EPYC 7742} processor and a graphics card \textit{NVIDIA A100}, the times were 240 and 752 minutes, respectively.
In contrast, the architecture designed in the FPGA enables the processing of these events in less than 0.2 seconds, assuming maximum throughput.

\section{Summary}
\label{sec:summary}
This paper presents the modified DIF algorithm for event stream filtration designed for an efficient FPGA implementation.
It achieved an AUROC of 0.907 for the diverse test sequences used, which is just slightly lower than the 0.913 value achieved by the well-known NNb method. Moreover, it also outperformed very recent methods based on deep neural networks.\remove{ The} Evaluation \replace{on}{for} a subset of test sequences showed that the DIF method scored 0.850 while EDnCNN 0.776 and AEDNet 0.844. The set was limited due to the huge computational complexity of the DNN methods. The proposed method is also characterised by a much smaller drop in\new{ filtering} effectiveness at high noise levels, which was 1\% on average, while for the NNb, EDnCNN, and AEDNet methods, they were 9\%, 5\%, and 14\%, respectively. 
We also proposed a hardware architecture for FPGA reconfigurable resources, which achieved high throughputs of \replace{\mbox{\SI{317.90}{MEPS}}}{\mbox{\SI{312.52}{MEPS}}} for \(1280 \times 720\) resolution and \replace{\mbox{\SI{408.56}{MEPS}}}{\mbox{\SI{400.26}{MEPS}}} for \(640 \times 480\) resolution. These throughputs are equal to \replace{\mbox{\SI{445.83}{MEPS}}}{\mbox{\SI{403.39}{MEPS}}} and \replace{\mbox{\SI{469.04}{MEPS}}}{\mbox{\SI{428.45}{MEPS}}}, respectively, in the version without global update of the inactive areas.
Additionally, a diverse dataset with a resolution of \mbox{\(1280 \times 720\)} has been prepared to evaluate the effectiveness of the filtering methods.

It is possible to add several optimisations to the proposed hardware architecture.
One would be to move the area activity information from the registers to BlockRAM. This change should result in a smaller decrease in throughput for an architecture with global updates.
Another optimisation could\new{ be} related to the storage of features in memory. As data is always read from four adjacent areas, it is not necessary for each of the four memories to contain all the information. It would be possible to divide the subareas so that each of the four neighbouring features is stored in separate memories, but not in the others. This would reduce the memory used by a factor of four, but the reading logic would be more complicated.
Another option would be to use one BlockRAM instead of four to store\new{ the} distance values for \(dx\) and \(dy\) with greater depth. This change could reduce resource utilisation and indirectly lead to a higher maximum operating frequency.
Additionally, since reading and writing data to and from the BlockRAMs that store timestamps and intervals \replace{is}{are} the same, they can be combined into four memories instead of eight, with twice the depth. This should have a positive effect on the amount of logical resources used, although the amount of BlockRAM used will remain the same.
\new{It is also planned to evaluate the proposed system using the \textit{Kria KV260 starter kit}, which includes the \textit{IMX636} sensor connected to the \textit{Zynq Ultrascale+ Multiprocessor System on Chip} (MPSoC) with the \textit{Petalinux} operating system. This will allow us to test the proposed algorithm in real-time operation.}
Another idea that could improve filtering efficiency is to use features other than the filtered timestamp and interval. Such features could also be calculated by a neural network (simple, spiking, or recurrent). However, this would require a careful hardware-aware design approach, as neural network models are usually quite computationally complex.

\section*{Acknowledgment}
The work presented in this paper was supported by the National Science Centre project no. 2021/41/N/ST6/03915 entitled ``Acceleration of processing event-based visual data with the use of heterogeneous, reprogrammable computing devices'' (first author) and the programme ``Excellence initiative – research university'' for the AGH University of Krakow (second author).
We gratefully acknowledge Polish high-performance computing infrastructure PLGrid (HPC Centers: ACK Cyfronet AGH, WCSS) for providing computer facilities and support within computational grant no. PLG/2023/016388.

\section*{Declaration of generative AI and AI-assisted technologies in the writing process}
During the preparation of this work, the authors used \textit{DeepL Write} in order to improve the readability. After using this tool, the authors reviewed and edited the manuscript as needed and take full responsibility for the content of the published article.


 \bibliographystyle{elsarticle-num} 
 \bibliography{cas-refs}





\end{document}